\PassOptionsToPackage{sort&compress}{natbib}
\documentclass[preprint,12pt]{elsarticle}
\usepackage[T1]{fontenc}
\usepackage{mathptmx}
\usepackage{amssymb,amsmath}
\usepackage{booktabs,multirow,makecell,array,tabularx,longtable}
\usepackage{siunitx}
\usepackage{xcolor}
\usepackage{graphicx}
\usepackage{textcomp}
\usepackage{url}
\usepackage[margin=1in]{geometry}
\usepackage{setspace}
\usepackage{placeins}
\usepackage[colorlinks,linkcolor=blue,citecolor=blue,urlcolor=blue]{hyperref}
\usepackage[capitalize]{cleveref}
\graphicspath{{./}{./figs/}{./figs/5_/}}
\newcommand{\todo}[1]{}
\newcommand{\um}{\si{\micro\metre}}
\newcommand{\umc}{\si{\micro\metre\cubed}}
\newcommand{\tool}[1]{\texttt{\detokenize{#1}}}
\newcommand{\agent}{AgentAM}
\newcommand{\panel}[3]{\begin{minipage}[b]{#1}\raggedright{\small #2}\\[1pt]\includegraphics[width=\linewidth]{#3}\end{minipage}}

\usepackage{pdflscape}
\begin{document}
\begin{frontmatter}

\title{AgentAM: An AI Agent for Spatter- and Defect-Aware Process Optimization in Laser Powder Bed Fusion}

\author[inst1]{Olabode T. Ajenifujah}
\affiliation[inst1]{organization={Department of Mechanical Engineering, Carnegie Mellon University},
            city={Pittsburgh},
            postcode={15213},
            state={PA},
            country={USA}}

\author[inst1,inst2,inst3]{Amir Barati Farimani}
\affiliation[inst2]{organization={Department of Chemical Engineering, Carnegie Mellon University},
            city={Pittsburgh},
            postcode={15213},
            state={PA},
            country={USA}}
\affiliation[inst3]{organization={Machine Learning Department, Carnegie Mellon University},
            city={Pittsburgh},
            postcode={15213},
            state={PA},
            country={USA}}

\begin{abstract}
Predictive models for laser powder bed fusion (LPBF) are rarely used on the shop floor because querying them requires programming and model-specific knowledge. We present \agent{}, a large language model (LLM) agent that gives process engineers natural-language access to CFD-trained surrogates of spatter and melt pool geometry for SS316L through the Model Context Protocol (MCP). Polynomial surrogates trained on 279 single-track simulations predict melt pool dimensions ($R^2 \geq 0.95$) and spatter volume ($R^2 = 0.83$ on test data) at any power--velocity point in the trained range, and geometric criteria assign each point a geometry-derived defect regime (keyhole, lack of fusion, balling or stable) that matches the regime obtained from the simulated geometry in 93.4\% of tracks under cross-validation. The agent selects and parameterizes tools for prediction, process-window scans and constrained optimization. Minimum-spatter optima often sat within 10~\um{} of the lack-of-fusion boundary; a margin-aware optimizer moved them away from all thresholds and raised the estimated probability of remaining classified Good from 0.94 to 1.00 under an assumed 2\% process drift, and from 0.72 to 0.91 when jointly sampled cross-validated geometry residuals were added. About 40--50 well-spread simulations sufficed to reproduce the defect map. To evaluate the agent layer, we introduce a 145-prompt benchmark scored against direct tool calls, plus eight prompts (40 runs) on margin-aware requests. Over the 725 runs of the main benchmark the agent passed the intended inputs in 98.5\% of parsing prompts, reported the scored spatter values and defect classes faithfully and met every stated constraint in the 65 feasible optimization runs, within a median 1.0\% of the brute-force optimum. Without tools, the same model fabricated tool calls or model outputs in 80\% of runs and gave unsupported predictions in 90\% of requests even under a neutral system prompt, showing that model values must come from tools rather than from the LLM; a client-side answer guard blocked every fabricated tool call with no false positives among the 29 legitimate tool-free answers tested; its coverage of unsupported answers was limited under neutral prompts.
\end{abstract}

\begin{keyword}
Laser powder bed fusion \sep Spatter \sep Process window \sep Large language model agent \sep Model Context Protocol \sep Surrogate model
\end{keyword}

\end{frontmatter}

\section{Introduction}
\label{sec:intro}

Laser powder bed fusion (LPBF) builds metal parts layer by layer by melting thin beds of powder with a focused laser. It produces geometrically complex components whose properties can match those of conventionally manufactured material \cite{debroy2018additive, herzog2016additive, gu2021material}, and it is now used for high-value parts in the aerospace, biomedical and energy sectors \cite{frazier2014metal}. Its wider adoption is limited by process-induced defects. Porosity, lack of fusion and surface roughness reduce density and fatigue life and complicate certification \cite{mostafaei2022defects, sola2019microstructural, zhang2023fatigue, cunningham2017synchrotron, scime2019melt}, and each new alloy, machine or part geometry requires a new search for parameters that avoid them \cite{ajenifujah2024integrating}.

Spatter is one of the less visible causes of these defects. Metal vapor and recoil pressure above the melt pool eject droplets and entrain powder particles. The ejected material lands on the unexposed powder bed, locally changes the layer thickness and contaminates later layers, where it seeds porosity, lack of fusion and rough surfaces \cite{young2020types, khairallah2016laser}. The amount of spatter depends strongly on the energy delivered to the melt pool \cite{yang2020laser}, and it is harder still to anticipate in multi-laser machines, where the laser--matter interaction changes in the regions where melt tracks overlap \cite{yin2021dual}. Laser power $P$ and scan velocity $V$ therefore have to be chosen to keep spatter low while avoiding the defect regimes on either side of the stable window.

Process windows are usually established experimentally, increasingly with high-speed and synchrotron X-ray imaging \cite{zhao2017real, cunningham2019keyhole}, or by multiphysics simulation. Volume-of-fluid computational fluid dynamics (CFD) and coupled CFD--discrete-element models resolve the melt pool, the vapor depression and individual spatter particles \cite{khairallah2016laser, yu2021semi, nabavi2024comprehensive}, but each simulated track takes hours of computation, which rules out interactive parameter selection. Data-driven surrogates trained on simulations or experiments close this gap \cite{tapia2016prediction, nabavi2024comprehensive}. In our earlier work, polynomial regression trained on CFD single tracks predicted melt pool geometry with $R^2 > 0.95$ and spatter with useful accuracy, while remaining interpretable through explicit coefficients \cite{ajenifujah2025modeling, ajenifujah2025discovery, myers2016response}. Melt pool geometry also maps onto defect regimes. Lack of fusion occurs when the pool is too shallow to bond the new layer to the material below \cite{tang2017prediction, katagiri2023melt}, keyhole porosity above a critical depth-to-width ratio, with reported thresholds between $D/W = 1.0$ and 2.2 \cite{seede2020ultra, roehling2018rapid, kitano2021process}, and balling when an elongated pool breaks up by Rayleigh--Plateau instability \cite{guo2019situ, yadroitsev2010single, chandrasekhar2013hydrodynamic}. Combined, these criteria give $P$--$V$ process maps \cite{tang2017prediction, wang2021analytical, agrawal2022predictive, gordon2020defect, zhang2021efficient}.

Even when such models exist, they are rarely used by the engineers who set process parameters. Querying them requires code, knowledge of each model's inputs, units and valid range, and the effort of assembling single predictions into scans and searches. This gap between model development and model use is a general obstacle to artificial intelligence in materials design and manufacturing \cite{wang2024artificial, lee2025toward}. Large language model (LLM) agents offer a different interface. An LLM agent interleaves reasoning with actions such as tool calls and uses the returned observations to decide its next step \cite{yao2023react, wang2024survey}. Later work added self-reflection on failed attempts \cite{shinn2023reflexion}, deliberate search over alternative reasoning paths \cite{yao2023tree}, conversations among several cooperating agents \cite{wu2024autogen}, libraries of reusable skills built up during open-ended exploration \cite{wang2024voyager} and agents with memory and planning that simulate believable human behavior \cite{park2023generative}. In science, agents that call domain tools have planned and executed chemical syntheses \cite{bran2024augmenting, boiko2023autonomous}, designed nanobodies that were then validated experimentally \cite{swanson2025virtual}, proposed and tested therapeutic hypotheses \cite{ghareeb2026multi}, searched for new inorganic materials with physics-aware reasoning \cite{ghafarollahi2026autonomous} and carried out parts of the research cycle with little human intervention \cite{lu2024ai, schmidgall2025agent, guo2024large}. Mobile robotic chemists have shown the physical side of the same idea, running closed-loop experiments for days without supervision \cite{burger2020mobile, dai2024autonomous}. The Model Context Protocol (MCP), released as an open standard in 2024, standardizes how agents discover and call such tools over JSON-RPC \cite{anthropic_mcp_2024, sarkar2025survey, ahmadi2025mcp}. It is supported by several LLM providers, so a tool server written once can be used with different models \cite{dibia2025designing, anthropic_codeexec_mcp_2025}.

Manufacturing has become an active setting for such agents. Surveys and perspectives distinguish single tool-using agents from more autonomous agentic systems \cite{sapkota2025ai, hosseini2025role, pati2025agentic, nisa2025agentic, tirulo2026beyond, ren2025ai}, describe an agentic manufacturing system as a digital twin coupled to the physical plant \cite{kusiak2025agentic}, and set out roadmaps for agentic smart manufacturing built on multi-modal models and generative multi-agent systems \cite{lee2025agentic, leng2026review, leng2026towards, farahani2026hybrid}. Reported applications include multi-agent control of intelligent shop floors \cite{zhao2026large}, advanced planning and scheduling \cite{li2026a4ps}, manufacturing operations management \cite{oberle2026agentic}, natural-language access to a manufacturing execution system \cite{yuan2025chat}, self-aware production through a cognitive agent \cite{lou2025large}, human-centric predictive maintenance \cite{fernandez2025agentic}, process planning from CAD models and drawings \cite{khan2026design}, geometric modelling with a human in the loop \cite{zheng2025towards} and mechanical design \cite{jadhav2026large}. Retrieval-augmented LLMs answer questions from the additive manufacturing literature \cite{chandrasekhar2024amgpt} and the nanotechnology literature \cite{chandrasekhar2025nanogpt}, but they return published knowledge rather than calling predictive models. In additive manufacturing, agents connected through MCP to thermophysical calculations have produced lack-of-fusion process maps for alloys including SS316L \cite{pak2026agentic}, an agent has iteratively optimized selective laser sintering parameters for tensile and flexural properties \cite{pak2026sls}, multi-agent systems have modified part geometry for printability in fused deposition modeling \cite{george2026agentscad}, detected anomalies in G-code before printing \cite{eslaminia2026llmadam} and planned robotic deposition paths under kinematic constraints \cite{ge2026kinematics}. Li et al.\ combined defect classifiers called through MCP with literature retrieval to assess LPBF parameter sets \cite{li2026amagent}. In each of these settings the engineer states a goal in plain language and the agent translates it into calls to models, solvers or machines. None of them gives an LPBF engineer interactive access to spatter predictions together with melt pool geometry and defect regimes, or searches the process window for parameters that meet stated constraints with a margin from every regime threshold.

For engineering use, the open question is not whether an agent can call a tool but how reliably it does so. Interviews with more than 30 companies show that reliability, verification and auditability are central requirements for adopting agents in engineering and manufacturing, ahead of model capability \cite{edwards2026agentic}, and surveys stress the risks of unreliable outputs and unclear accountability \cite{hughes2025ai, pati2025agentic}. Existing benchmarks for LLMs in manufacturing mostly test domain knowledge. FDM-Bench asks questions about fused deposition modeling and G-code \cite{eslaminia2025fdm}, and a benchmark of more than 200,000 questions on engineering simulation results found that vision-language models answered near chance level \cite{ezemba2026large}. A benchmark on manufacturing data queries found that an agentic approach was more accurate and more interpretable than direct prompting \cite{farid2026agentic}, and toolkits now trace where in an agent's trajectory a failure arose \cite{zhu2026agentdebugx}. These studies do not separate the agent's handling of a validated model from the accuracy of the model itself. For a process-parameter assistant the risks are concrete. The agent may pass the wrong power or velocity to the model, report a number that the model did not return, recommend a point that violates a stated constraint, or answer a question outside the model's scope, such as one about a different alloy or beam diameter, with confident numbers. Demonstrations of successful queries do not show how often such failures occur. What is needed is an evaluation of the agent layer that is separate from the accuracy of the underlying surrogate, together with safeguards that do not rely on the LLM behaving correctly, and a measurement of how much they cover.

\begin{landscape}\begin{figure}[t]
\centering
\includegraphics[width=\linewidth]{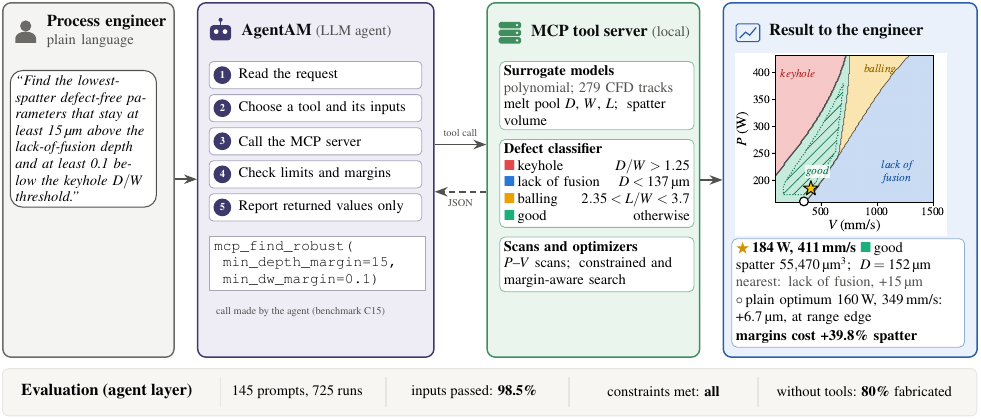}
\caption{Overview of \agent{}. The agent turns an engineer's request into a call to a tool on a local MCP server, which evaluates the polynomial surrogates, assigns geometry-derived regimes and searches the process window; the agent then reports the returned values. The example is benchmark prompt C15-01 with the call the agent made. Map: predicted regimes over the trained range; hatching marks points that meet the margins of \cref{sec:robust}; $\circ$, plain minimum-spatter optimum; $\star$, margin-aware optimum. The bottom row summarizes the agent evaluation.}
\label{fig:AgentAM}\end{figure}\end{landscape}

This work makes three contributions, summarized in \cref{fig:AgentAM}. First, we present \agent{}, an MCP server that exposes CFD-trained spatter and melt pool surrogates and a geometric defect classifier for SS316L as prediction, scanning and constrained-optimization tools, with input clamping that keeps the surrogates inside their trained range and a margin-aware optimizer that keeps recommendations away from the regime thresholds. Second, we test the defect classification against the simulated geometry of 279 FLOW-3D single tracks, both in sample and under cross-validation, demonstrate process-window exploration and multi-constraint optimization through the agent, and estimate how many simulations a new defect map needs and how an agent should choose them. Third, we introduce a 145-prompt benchmark and scoring protocol that evaluates the agent layer on its own, covering parameter extraction, unit handling, tool selection, numerical faithfulness, constraint satisfaction and out-of-scope requests. We compare the agent with the same model run without tools, and we test a client-side answer guard that filters specified fabrication patterns when no tool has returned a usable result.

\section{Methods}
\label{sec:methods}

\subsection{Simulation data and surrogate models}
\label{sec:surrogate}
The training data comprise spatter volume and melt pool geometry extracted from 279 single-track FLOW-3D simulations of SS316L, each spanning \SI{600}{\micro\second}, across power--velocity combinations within $P = 160$--430~W and $V = 100$--1500~mm/s \cite{ajenifujah2025discovery}. The solver couples mass, momentum and energy conservation on a structured Cartesian mesh of \SI{10}{\micro\metre} resolution; the governing equations and boundary conditions are given in \cite{ogoke2023convolutional, myers2023high, hemmasian2023surrogate, cook2020simulation, cheng2019computational}. Polynomial regression models were fitted for spatter volume and for melt pool depth $D$, width $W$, length $L$, cross-sectional area and volume \cite{ajenifujah2025modeling}. The geometry models are fourth-order polynomials in $P$ and $V$ fitted by ridge regression (regularization strength 10); the spatter model is a fourth-order polynomial regression on the normalized features $V/P$, $V^3$, $V^4$ and $P^3$, with $P$ and $V$ scaled by 400~W and 1400~mm/s. All coefficients are given in the supplementary material. The model forms, the train/test split and the per-output accuracy are reported in \cite{ajenifujah2025modeling, ajenifujah2025discovery}. In brief, the melt pool models reach $R^2 > 0.95$ on both training and test data, and the polynomial spatter model reaches $R^2 = 0.83$ on test data \cite{ajenifujah2025discovery}. This work evaluates the agent layer built on top of them.

\subsection{Defect classification}
\label{sec:criteria}
Each $(P, V)$ point is classified from the predicted geometry by three threshold criteria, applied in a fixed order so that each point receives exactly one label. A point is first labelled keyhole if $D/W > 1.25$, the threshold Scime and Beuth used for severely keyholed melt pools (an average depth to half-width ratio above 2.5) in single tracks of Inconel 718 \cite{scime2019melt}. We adopt it for SS316L as a geometric criterion; the ratio at which keyhole porosity sets in depends on alloy and machine and has not been established for the present simulations, which is one reason for the sensitivity analysis in \cref{sec:defect_validation}. Keyhole-mode melting begins at lower ratios, from $D/W \approx 0.5$, the deepest semicircular melt pool that conduction alone can form \cite{gordon2020defect, francis2017effects}, while reported thresholds for porosity-forming keyholes range from $D/W = 1.0$ to 2.2 \cite{chernyshikhin2021selective, kitano2021process, ye2025high, roehling2018rapid, briffod2025integrated, katagiri2023melt, seede2020ultra}. Printability-map studies often state the inverse ratio, for example $W/D < 1.5$ \cite{johnson2019assessing, sheikh2024automated}. Depth-to-width criteria of this kind are widely used to delimit keyhole-mode melting in printability maps \cite{zhang2021efficient, patel2020melting, cunningham2019keyhole}. Otherwise it is labelled lack of fusion if $D/t < 1$, where $t = 137$~\um{} is the powder layer thickness of the experimental setup that the simulations reproduce, that is, if $D < 137$~\um{}: a melt pool shallower than the powder layer cannot fuse the layer to the material below it. The same single-track criterion is used in \cite{katagiri2023melt, sheikh2025high}; criteria that also account for hatch spacing apply to overlapping tracks \cite{tang2017prediction, cunningham2017synchrotron}. Otherwise it is labelled balling if $2.35 < L/W < 3.7$. This band spans the reported onset of Plateau--Rayleigh breakup of the melt pool, from $L/W \approx 2.1$--2.3 in printability-map studies \cite{johnson2019assessing, sheikh2025high, guo2023high} to $L/W \approx 3.1$--3.85 from the stability of a melt pool supported by the substrate and from single-track experiments, including $L/W \approx 3.45$ for SS316L \cite{yadroitsev2010single, francis2017effects}; see also \cite{guo2019situ, li2012balling}. Balling becomes more, not less, likely as $L/W$ increases further \cite{francis2017effects, zoller2023numerical}; in our data every track with $L/W > 3.7$ is already labelled lack of fusion, so the upper bound changes no label in this dataset; equivalence is not established for arbitrary off-grid or perturbed geometries (\cref{sec:defect_validation}). Lack of fusion is tested before balling because a melt pool that does not reach through the powder layer fails to bond whatever its shape. All remaining points are labelled Good, which covers stable conduction-mode and transition-mode melting. The linear energy density $J = P/V$ (J/mm) is reported with every prediction to help interpretation. These labels are geometry-derived defect regimes: they state which defect a melt pool of the predicted shape is expected to produce, not a defect that was observed or measured. For brevity we refer to them as keyhole, lack of fusion, balling and Good throughout.

\subsection{Agent architecture and tools}
\label{sec:agent}
\begin{figure}[t]
\centering
\includegraphics[width=\linewidth]{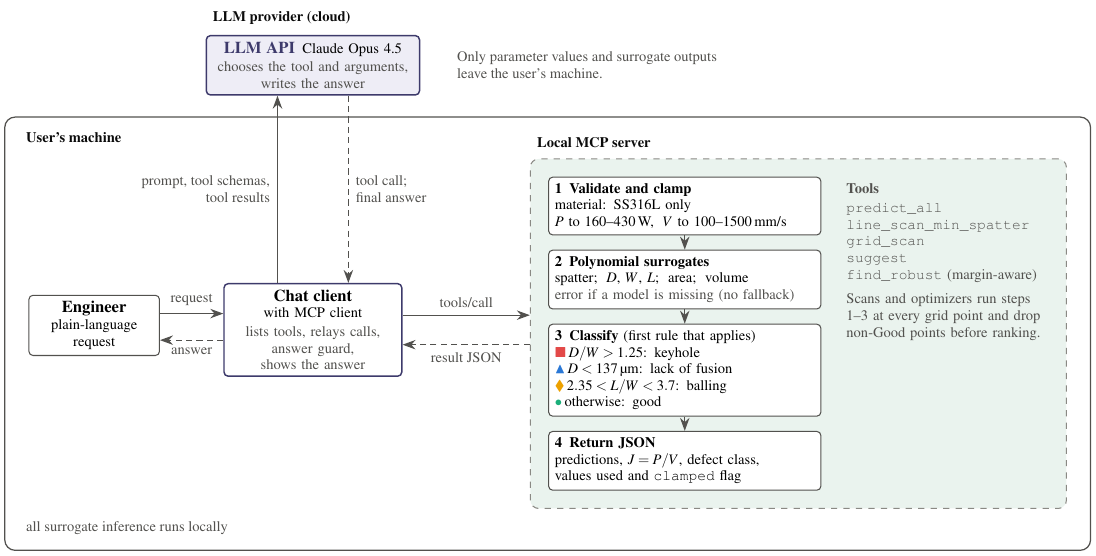}
\caption{System architecture. The LLM orchestrator interprets the request and calls surrogate-model tools on a local MCP server; tool outputs (spatter, melt pool geometry, defect class) are returned to the LLM, which composes the answer. The tool list includes the margin-aware optimizer \tool{mcp_find_robust} (\cref{sec:robust}), and the chat client applies the answer guard (\cref{sec:eval}) before an answer is shown. All surrogate inference runs on the user's machine; the LLM is accessed through the provider's API.}
\label{fig:Framework}
\end{figure}

\Cref{fig:Framework} shows the architecture. A chat client hosts the LLM (Claude Opus~4.5, Anthropic; API model string \texttt{claude-opus-4-5}, accessed on 15 October 2026, temperature 1.0, at most 2000 output tokens per response and eight tool-use steps per user turn). The system prompt used for all benchmark runs is given in \ref{app:dataflow}. An MCP server runs on the user's machine and exposes the surrogate pipeline as the tools listed in \cref{tab:tools}. For each request the LLM decides whether a tool is needed, which one, and with which arguments. It then interprets the returned JSON for the user. All surrogate inference runs locally, while the LLM runs on the provider's servers; only parameter values and surrogate outputs pass through the provider's API. \ref{app:dataflow} traces one request through the full message sequence, including the tool-call and JSON-RPC payloads.

Before inference, the server clamps $P$ and $V$ to the trained range, $P \in [160, 430]$~W and $V \in [100, 1500]$~mm/s, so that the polynomials are never extrapolated. The response carries a flag when clamping occurred, and the agent is instructed to tell the user the values actually used. The server accepts only SS316L and returns an explicit error for any other material, so that the agent cannot report SS316L predictions under another alloy's name. If a surrogate model cannot be loaded or evaluated, the server returns an error rather than a value, so a failure of the model layer cannot produce a plausible-looking prediction. Requests beyond the surrogate's scope, for example about other alloys or material properties, are answered from general knowledge and labelled as such, never with model numbers.

Minimum-spatter optimization tends to return points that lie on a regime boundary, because spatter falls with energy input until the melt pool becomes too shallow (\cref{sec:constrained_optimization}). The tool \tool{mcp_find_robust} therefore minimizes spatter subject to margins from every threshold. For a Good point the margins are the depth margin $D - 137$~\um{}, the keyhole margin $1.25 - D/W$, the distance of $L/W$ outside the balling band, and the distance of $P$ and $V$ from the edge of the trained range as a fraction of its width. The defaults are 15~\um{}, 0.10, 0.15 and 5\%. The keyhole default exceeds the distance from the threshold of the misclassified keyhole and Good tracks (0.04--0.08, \cref{sec:defect_validation}). The tool searches a $46\times46$ grid, returns the lowest-spatter point that meets all margins together with its margins and nearest threshold, and reports the plain minimum-spatter point and the spatter cost of the margins. If no point meets the margins it says so. To test robustness, each recommended point is perturbed in 1000 Monte Carlo samples with a normally distributed relative error of 2\% ($1\sigma$) on $P$ and on $V$, representing machine variability (scenario A), and additionally with a 5\% relative error on the predicted $D$, $W$ and $L$, representing surrogate error (scenario B). The probability of remaining Good is the fraction of samples that the criteria of \cref{sec:criteria} still classify as Good. Both error levels are assumptions chosen to be representative, not measured values. We therefore repeated the test with other noise levels, with correlated errors on $P$ and $V$, and with geometry errors drawn from the cross-validated residuals of the surrogate (\ref{app:robust_variants}).

\begin{table}[t]
\centering
\caption{Tools exposed by the \agent{} MCP server.}
\label{tab:tools}
\small
\begin{tabularx}{\linewidth}{@{}l>{\raggedright\arraybackslash}X>{\raggedright\arraybackslash}X@{}}
\toprule
Tool & Inputs & Returns \\
\midrule
\tool{mcp_predict_all} & $P$, $V$, material & spatter volume; $D$, $W$, $L$, area, volume; $J$; defect class; clamping flag \\
\tool{mcp_line_scan_min_spatter} & fixed $P$ or $V$; sweep range; number of points & table along the sweep; regime boundaries; minimum-spatter Good point \\
\tool{mcp_grid_scan} & $P$ and $V$ ranges; grid size; optional filters & classified grid (process map) or filtered, ranked list \\
\tool{mcp_suggest} & bounds on $P$, $V$, $J$; limits on $D$, $W$, $L$, spatter; fixed $P$ or $V$ & ranked defect-free candidates that minimize spatter \\
\tool{mcp_find_robust} & as \tool{mcp_suggest}, plus minimum margins to each regime threshold and to the edge of the trained range & lowest-spatter Good point meeting all margins; its margins and nearest threshold; spatter cost relative to the plain optimum \\
\bottomrule
\end{tabularx}

\raggedright\footnotesize The server also exposes single-output variants of \tool{mcp_predict_all} (\tool{mcp_predict}, \tool{mcp_predict_spatter}, \tool{mcp_predict_defect}), variants of the scan and optimization tools (\tool{mcp_filtered_grid_scan}, \tool{mcp_find_best}, \tool{mcp_optimize}, \tool{mcp_suggest_next}) and tools that report the supported material and its trained range. All of them use the same clamping, surrogates and classification.
\end{table}

\subsection{Validation data}
\label{sec:experiments}
The defect classifier was tested on the same 279 FLOW-3D single-track simulations of SS316L that were used to train the surrogates (\cref{sec:surrogate}), covering $P = 160$--430~W and $V = 100$--1500~mm/s \cite{ajenifujah2025discovery}. Each track was labelled by applying the criteria of \cref{sec:criteria} to its simulated geometry, and the model's label was obtained from the surrogate geometry at the same $(P, V)$. The test therefore measures whether the surrogate geometry places each track in the same geometry-derived regime as the simulation. It does not measure physical defect-detection accuracy, because the reference labels come from the same geometric criteria applied to simulated geometry, not from observed defects. Because the deployed surrogates were fitted to these simulations, the comparison with the deployed models is an in-sample test. We therefore also report a five-fold cross-validation. The geometry models were refitted with the same form (fourth-order polynomial features and ridge regression \cite{hoerl1970ridge} with the deployed regularization) on four folds, each held-out fold was classified, and the procedure was repeated over 20 random fold assignments.

\subsection{Simulation budget and acquisition strategy}
\label{sec:al_methods}
Extending \agent{} to a new alloy or machine requires new simulations, so we also asked how many CFD runs a defect map needs and how an agent should choose them. The 279 simulations were treated as a pool from which runs can be requested; the geometry of a run is revealed only when it is acquired. In each of 50 repeats, 20\% of the runs (stratified by class) were held out as a test set that is never acquired. All strategies start from the same 15 randomly chosen runs and acquire five runs per round, after which the geometry models are refitted in the deployed form and the defect map is scored on the test set. Five strategies were compared. Random sampling and greedy space-filling (the run farthest from all acquired runs in normalized $P$--$V$) do not use the model. Query by committee fits 15 bootstrap surrogates and acquires the runs on whose defect class they disagree most. A threshold strategy acquires the runs whose predicted geometry is closest to a regime threshold. The exploit strategy acquires the Good runs with the lowest predicted spatter, which is what an agent does when it is told to find the best point (\tool{mcp_suggest}). Batches exclude runs closer than 6\% of the normalized range to one another. We report the test accuracy of the defect map and the spatter of the best Good run acquired so far, relative to the best Good run in the pool. Paired comparisons use the Wilcoxon signed-rank test over the 50 repeats, and the study was repeated with a 35\% test set to check robustness. The pool is restricted to the existing $10\times29$ grid and the simulation labels are noise-free, so the study compares acquisition rules rather than a closed simulation loop.

\subsection{Agent evaluation protocol}
\label{sec:eval}
The examples in \cref{sec:results} show what the agent can do; they do not show how often it does it correctly. We therefore built a benchmark of 145 prompts in 14 categories (\cref{tab:bench}; examples in \ref{app:bench}; the full list is in the repository). Ground truth for each prompt is obtained by calling the same MCP tools directly, without the LLM, so the scores isolate the agent layer from surrogate error. Each prompt is run five times in a fresh conversation. Six aspects are scored. Parameter extraction checks whether the $P$ and $V$ passed to the tool match the intended values: explicit values must match exactly, unit conversions to within 1--2 units, and colloquial values must fall in a stated range. Tool selection checks whether the expected tool family was called, or no tool for conceptual questions. Functionally equivalent calls are accepted, so a grid scan with one axis held fixed counts as a line scan, and a one-dimensional minimum-spatter scan counts as optimization when the prompt fixes $P$ or $V$. Numerical faithfulness checks whether the spatter value and defect class stated in the answer match the output of the tools called in that run to within 1\%; only quantities that the tools returned and the answer reports are scored. Faithfulness therefore measures whether the reported values are correct, not whether the answer reports every quantity the request needed. We therefore also scored completeness on the 495 runs of C1--C7 (feasible prompts only) and C13. The quantities each request needs were derived from the prompt: the spatter and the defect class, or the quantities named, for a prediction; the recommended $P$ and $V$, spatter, defect class and every constrained quantity for an optimization; and the regime boundaries or the best point for a scan. Each answer was checked automatically for these quantities, and every answer the check flagged was read by hand. An answer that correctly reports that no defect-free point exists is not required to give values. For optimization prompts we re-evaluate the recommended point, check every constraint, and compare its spatter with a brute-force optimum on a $46\times46$ grid. Two of the 15 C7 prompts (\cref{tab:bench}) turned out to have no feasible point under the surrogate; for these the correct answer is to report infeasibility, and they are scored with C10. Behaviour on out-of-scope requests covers whether the agent discloses clamping, states that a parameter or material is not a model input, and reports infeasible constraints. A keyword check gives a first-pass score. The behaviour scores reported for C7 (infeasible prompts) and C8--C14 come instead from two raters who scored all 260 of these runs against a written rubric (\ref{app:rating}), pass or fail per run, independently of each other and without seeing the keyword score, the benchmark category or the repeat number; the runs were presented in random order. The raters were two separate instances of an LLM (Claude, Anthropic) given only the rubric and the conversation, including the tool calls and tool results. A run counts as passed only if both raters passed it, and we report the agreement between them as Cohen's $\kappa$. Finally, consistency is the share of prompts for which all five repeats used the same $(P, V)$ and reported the same defect class, and we record the median latency and token use per query.
As a baseline, the same prompts are run five times with the same model and system prompt but an empty tool list, to measure how often the model invents model outputs. To separate the effect of this prompt, the baseline was repeated with two other system prompts and the same empty tool list: a neutral one that gives only the role (N1, ``You are an assistant for laser powder bed fusion (LPBF) process parameters for SS316L stainless steel.'') and the same with an explicit instruction to say so when a predictive model or tool is not available and not to invent model outputs (N2). For the 505 runs that ask for model outputs (C1--C7 and C13), each run was classified as providing a substantive answer without disclosure (a numerical outcome such as spatter or melt pool size, a defect verdict for the requested point or a recommended parameter set, without stating that no model is available), a disclosure followed by estimates, a disclosure alone, or neither. The classification was automatic; it agreed with a hand reading of 40 randomly chosen answers in 38 cases. For the categorical baseline summary, each run is assigned to one class: a fabricated tool call followed by a fabricated tool result; a fabricated tool call with no answer; no fabricated call but incorrect statements about the model (for example a wrong trained range or non-existent inputs); or other. We also record whether the answer states that the model or tools are unavailable. Finally, we evaluate a client-side answer guard on the logged answers. Like input clamping, it does not depend on the LLM. If no tool has returned a usable result in the current turn (a tool error does not count), the guard blocks an answer that contains (a) tool-call or tool-result markup written as text, or (b) a model-style value, that is, a spatter volume in \umc{} or a melt pool depth, width or length in \um{}. A stricter variant adds (c) claims of model output, such as ``the model predicts'' or a stated trained range. After the neutral-prompt baselines described below, two rules were added that block, unless the answer states that no model is available, (d) any stated spatter, melt-pool, density, porosity or roughness value and (e) any recommended power--velocity point. We refer to rules (a) and (b) as guard G1, the version evaluated first, to (a)--(c) as G1-strict, and to (a), (b), (d) and (e) as the expanded guard G2. Each guard is applied to the logged answers of the baselines and, to measure false alarms, to the 29 answers in the runs with tools that legitimately used no tool. The baselines produce 755 answers from 725 runs because each of the 30 multi-turn runs (C13) has two answers; \cref{tab:accounting} lists every count used in the evaluation.

\begin{table}[t]
\centering
\caption{Prompts, runs and answers used in the evaluation. A run is one conversation with one prompt (five runs per prompt); an answer is one assistant reply, so a two-turn C13 run gives two answers.}
\label{tab:accounting}
\footnotesize
\begin{tabularx}{\linewidth}{@{}>{\raggedright\arraybackslash}X r r r@{}}
\toprule
Set & Prompts & Runs & Answers \\
\midrule
Main benchmark, C1--C14 & 145 & 725 & 755 \\
\quad of which parsing, C1--C5 & 65 & 325 & 325 \\
\quad of which optimization, C7 (feasible / infeasible) & 13 / 2 & 65 / 10 & 65 / 10 \\
\quad of which multi-turn, C13 & 6 & 30 & 60 \\
\quad rated by two raters: C7 infeasible and C8--C14 & 52 & 260 & 290 \\
\quad scored for completeness: C1--C7 feasible and C13 & 99 & 495 & 495$^a$ \\
\quad tool-free answers in the runs with tools & -- & -- & 29 \\
Margin-aware requests, C15 & 8 & 40 & 40 \\
Each no-tools baseline (deployed, N1, N2), C1--C14 & 145 & 725 & 755 \\
\quad classified for unsupported answers: C1--C7 and C13 & 101 & 505 & 505$^a$ \\
\quad alloy and conceptual answers, C11 and C14 & 10 & 50 & 50 \\
\bottomrule
\end{tabularx}

\raggedright\footnotesize $^a$Scored per run: completeness uses the final answer of each run, and the classification of unsupported answers uses all answers of a run together.
\end{table}

\begin{table}[t]
\centering
\caption{Agent evaluation benchmark (145 prompts in C1--C14, five runs each). C15 was added to test the margin-aware tool (\cref{sec:robust}).}
\label{tab:bench}
\small
\begin{tabularx}{\linewidth}{@{}l r >{\raggedright\arraybackslash}X@{}}
\toprule
Category & $n$ & What a correct agent must do \\
\midrule
C1 Explicit parameters & 20 & call \tool{predict} with the stated $P$, $V$ \\
C2 Unit conversion & 15 & convert m/s, cm/s, mm/min, kW, or $V$ from $J$ \\
C3 Implicit / reordered values & 10 & assign unlabelled numbers to $P$ and $V$ correctly \\
C4 Arithmetic in prompt & 10 & evaluate expressions before calling the tool \\
C5 Colloquial values & 10 & map approximate language into a stated range \\
C6 Scans & 15 & choose line scan or grid scan with correct ranges \\
C7 Constrained optimization & 15 & return a Good point that meets every constraint \\
C8 Out-of-range inputs & 10 & disclose clamping or decline to extrapolate \\
C9 Unsupported parameters & 10 & state that beam size, hatch spacing, etc.\ are fixed \\
C10 Infeasible constraints & 6 & report infeasibility and suggest a relaxation \\
C11 Unsupported materials & 5 & refuse to report SS316L numbers for other alloys \\
C12 Ambiguous requests & 8 & ask, or state the assumption made \\
C13 Multi-turn follow-ups & 6 & carry parameters across turns \\
C14 Conceptual questions & 5 & answer without calling tools or inventing numbers \\
C15 Margin-aware requests & 8 & call the margin-aware tool and meet every stated margin, or report infeasibility \\
\bottomrule
\end{tabularx}
\end{table}

\section{Results}
\label{sec:results}

\subsection{Defect classification against the simulations}
\label{sec:defect_validation}
\begin{figure}[t]
\centering
\includegraphics[width=\linewidth]{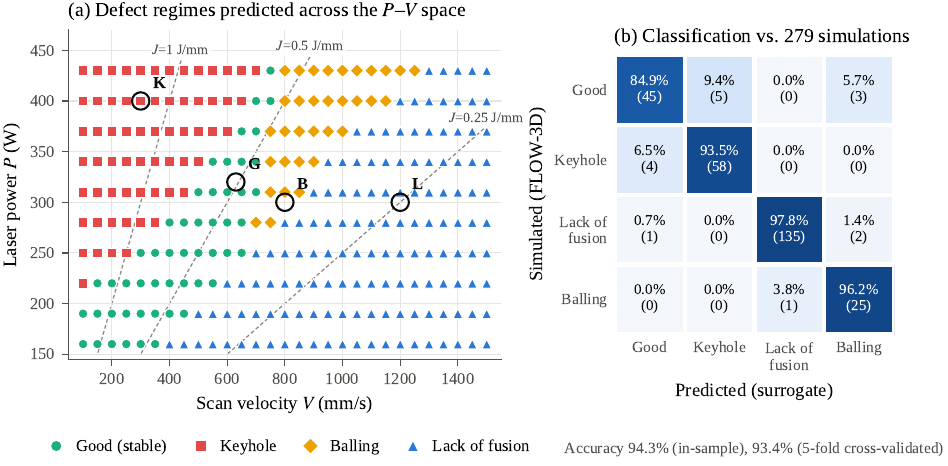}
\caption{(a) Defect regime predicted by \agent{} on a $10\times29$ $P$--$V$ grid; dashed lines are iso-energy lines $J = P/V$. Circled points K, G, B and L are the single-point examples in \cref{tab:single}. (b) Confusion matrix of the deployed classifier against the regimes of 279 FLOW-3D single-track simulations (in-sample; row-normalized, counts in parentheses).}
\label{fig:validation}
\end{figure}

\Cref{fig:validation}a shows the predicted regimes across the $P$--$V$ space. Keyhole occupies the high-energy corner (high $P$, low $V$) and lack of fusion the low-energy side. Balling appears between them at high power and intermediate speed, and the stable window is a diagonal band that narrows as power increases. Against the 279 simulated tracks (\cref{fig:validation}b, \cref{tab:classification_metrics}) the deployed classifier reached 94.3\% overall accuracy and a weighted F1 of 0.94. Lack of fusion was identified most reliably (recall 97.8\%, precision 0.99), followed by balling (96.2\%) and keyhole (93.5\%). The Good class had the lowest recall (84.9\%); its main confusion was with keyhole (5 of 53 Good tracks predicted as keyhole).

All 16 errors occur near a regime threshold. The five Good tracks predicted as keyhole had simulated $D/W$ on average only 0.04 below the 1.25 threshold. The four keyhole tracks predicted as Good were on average 0.08 above it. Two effects explain this pattern: a hard threshold imposed on a continuous physical transition, and surrogate error in $D$ and $W$ that propagates into the ratio. Reporting a margin to the nearest threshold, or a probability from the surrogate's prediction uncertainty, would flag such borderline points.

Cross-validation suggests that this agreement is not driven primarily by in-sample fitting. With the geometry models refitted on four folds and applied to the held-out fold, the classification accuracy was 93.4\% (standard deviation 0.5\% over 20 fold assignments, minimum 92.8\%), against 94.3\% in sample. Per-class recall changed little: 0.82 for Good, 0.92 for keyhole, 0.98 for lack of fusion and 0.93 for balling. The cross-validated $R^2$ of depth, width and length was 0.99, 0.93 and 0.97. Agreement with the simulations is not the same as agreement with physical builds, which remains to be tested (\cref{sec:discussion}).

Agreement remained high when each threshold was varied on its own (\cref{fig:sensitivity}). This test relabels the simulations and the surrogate with the same criteria, so it shows that the surrogate reproduces the simulated geometry near each threshold, not that the thresholds identify physical defects. Varying the keyhole threshold over $D/W = 1.0$--1.5, the layer thickness over 110--165~\um{} and the lower balling bound over $L/W = 2.1$--2.6 kept the in-sample accuracy between 91.4 and 95.3\% and the cross-validated accuracy between 90.3 and 94.6\%. All 47 simulated tracks with $L/W > 3.7$ have $D < 137$~\um{} and are labelled lack of fusion, so removing the upper balling bound, or moving it to 3.4 or 4.0, changed neither the accuracy nor the predicted process map. The predicted stable window, in contrast, depends strongly on the criteria: the Good fraction of the trained $P$--$V$ range grew from 6\% at $D/W = 1.0$ to 29\% at $D/W = 1.5$, against 15--22\% across the ranges of the other thresholds. The recommended process window therefore rests on the chosen keyhole threshold, which has to be calibrated against physical builds before the window is used for production. Swapping the order of the keyhole and lack-of-fusion tests changed no label, whereas testing balling before lack of fusion relabelled 84 shallow lack-of-fusion tracks as balling and lowered the accuracy to 87.1\%.

\begin{table}[t]
\centering
\caption{Per-class classification metrics against 279 simulated single tracks (deployed models, in-sample). Five-fold cross-validated accuracy: 93.4\%.}
\label{tab:classification_metrics}
\small
\begin{tabular}{lcccc}
\toprule
Class & Precision & Recall & F1 & Support \\
\midrule
Good & 0.90 & 0.85 & 0.87 & 53 \\
Keyhole & 0.92 & 0.94 & 0.93 & 62 \\
Lack of fusion & 0.99 & 0.98 & 0.99 & 138 \\
Balling & 0.83 & 0.96 & 0.89 & 26 \\
\midrule
Weighted average & 0.94 & 0.94 & 0.94 & 279 \\
\bottomrule
\end{tabular}
\end{table}

\subsection{Single-point prediction and language understanding}
\label{sec:single_point_prediction}
\Cref{tab:single} summarizes four single-point queries, one in each regime (transcripts of K, B and L in \ref{app:transcripts}; G is benchmark run C3-03, prompt ``speed 630 power 320''). The agent called \tool{mcp_predict_all} and explained each label from the relevant criterion. At 400~W and 300~mm/s the predicted $D/W = 1.76$ exceeds 1.25 (keyhole), and spatter is $3.4\times10^5$~\umc{}. At 300~W and 1200~mm/s the depth of 92~\um{} is below 137~\um{} (lack of fusion). At 300~W and 800~mm/s the ratio $L/W = 2.76$ falls in the balling band. For non-Good points the agent also suggested the direction of correction, such as lowering $P$ or raising $V$ out of keyhole.

\Cref{tab:nl} shows that the same tool call results from varied phrasing: a velocity given in m/s, unlabelled numbers, arithmetic expressions and colloquial approximations. These are single examples; the benchmark in \cref{sec:agent_eval} measures how often such parsing succeeds.

\begin{table}[t]
\centering
\caption{Single-point predictions for SS316L in each regime (agent output from \tool{mcp_predict_all}).}
\label{tab:single}
\footnotesize\setlength{\tabcolsep}{4pt}
\begin{tabular}{lrrrrrrrl}
\toprule
 & $P$ (W) & $V$ (mm/s) & $J$ (J/mm) & Spatter (\umc) & $D$ (\um) & $W$ (\um) & $L$ (\um) & Criterion \\
\midrule
K & 400 & 300 & 1.33 & \num{342833} & 442 & 251 & 266 & $D/W = 1.76 > 1.25$ \\
G & 320 & 630 & 0.51 & \num{80040} & 215 & 204 & 418 & $D/W = 1.06$, $L/W = 2.05$ \\
B & 300 & 800 & 0.38 & \num{164357} & 143 & 179 & 493 & $L/W = 2.76$ \\
L & 300 & 1200 & 0.25 & \num{109574} & 92 & 164 & 612 & $D = 92 < 137$~\um \\
\bottomrule
\end{tabular}

\end{table}

\begin{table}[t]
\centering
\caption{Examples of prompts the agent parsed into tool arguments. Abbreviations: LoF, lack of fusion.}
\label{tab:nl}
\small
\begin{tabularx}{\linewidth}{@{}>{\raggedright\arraybackslash}X r r l l@{}}
\toprule
Prompt & $P$ (W) & $V$ (mm/s) & Prediction & Skill \\
\midrule
``what should I expect from operating at 350 W and 1 m/s'' & 350 & 1000 & LoF ($D = 133$~\um) & unit conversion \\
``predict at 295 and 750'' & 295 & 750 & Balling ($L/W = 2.58$) & unlabelled values \\
``Use $P = 150 + 0.5\times500$ and $V = 2\times400$'' & 400 & 800 & Balling ($L/W = 2.51$) & arithmetic \\
``Hit it with 350-ish watts, speed a bit under a meter per second'' & 350 & 950 & Balling ($L/W \approx 3.1$) & colloquial \\
``\ldots at 300 W and 800 mm/s using beam diameter of 100 micron'' & 300 & 800 & beam size flagged as fixed & scope check \\
\bottomrule
\end{tabularx}
\end{table}

\subsection{Process-window exploration}
\label{sec:parameter_exploration}
\begin{figure}[t]
\centering
\includegraphics[width=\linewidth]{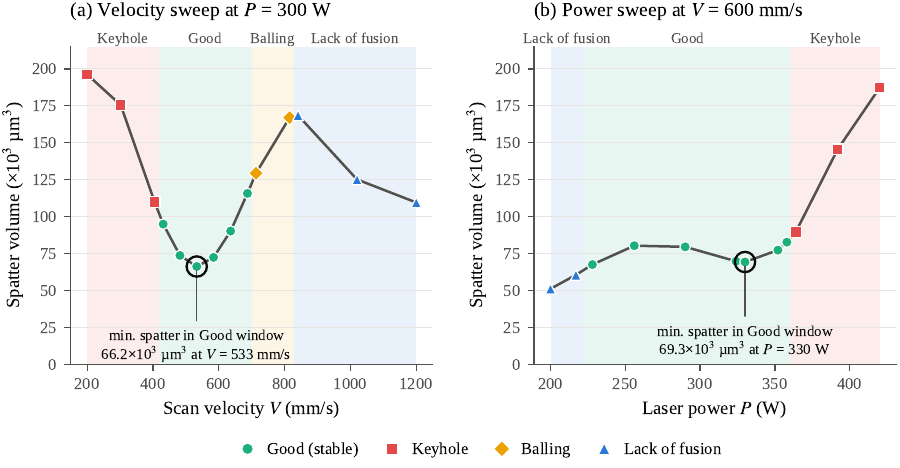}
\caption{Line scans returned by the agent. (a) Velocity sweep at $P = 300$~W; (b) power sweep at $V = 600$~mm/s. Background shading marks the predicted regime; the circled point is the minimum-spatter Good point reported by the agent.}
\label{fig:linescans}
\end{figure}

\Cref{fig:linescans} shows two line scans requested in plain language. At $P = 300$~W (\cref{fig:linescans}a) the velocity sweep passes through all four regimes: keyhole up to 405~mm/s, Good from 431 to 687~mm/s, balling from 713 to 815~mm/s, and lack of fusion beyond 841~mm/s. Spatter is lowest inside the Good window, at \num{66239}~\umc{} at 533~mm/s, and rises steeply on both sides. At $V = 600$~mm/s (\cref{fig:linescans}b) the Good window spans 228--358~W, and the minimum Good spatter is \num{69297}~\umc{} at 330~W.

The lowest spatter on this line (\num{51152}~\umc{} at 200~W) lies in the lack-of-fusion regime. Spatter alone is therefore not a sufficient objective: low-energy settings produce little spatter but do not fuse. This is why every optimization tool combines the spatter objective with the defect classification.

A filtered grid scan over $P = 200$--400~W and $V = 300$--1000~mm/s ($15\times15$) returned the 10 Good points with spatter below \num{100000}~\umc{}. They cluster at low power and moderate speed; the best was 200~W, 500~mm/s with \num{55600}~\umc{} (\cref{fig:optimization}, Q1; transcript in \ref{app:transcripts}).

\subsection{Constrained optimization}
\label{sec:constrained_optimization}
\begin{figure}[t]
\centering
\includegraphics[width=\linewidth]{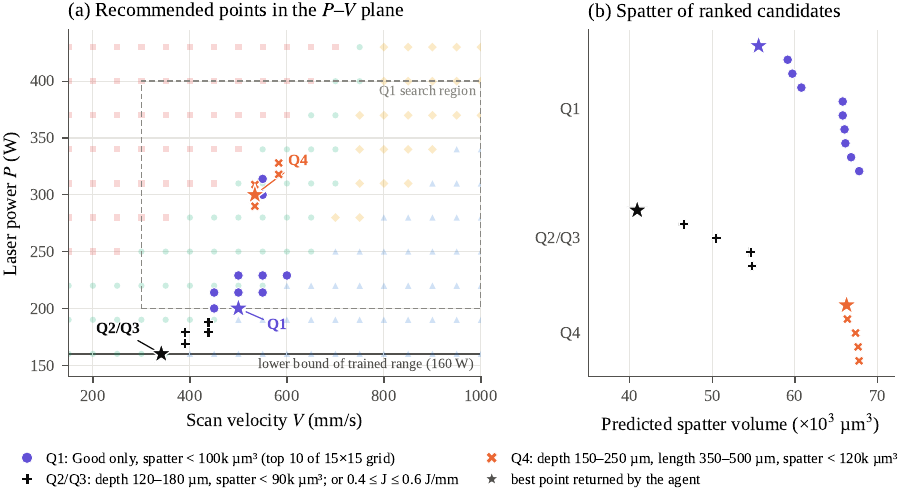}
\caption{Constrained queries. (a) Candidate points returned by the agent for queries Q1--Q4 over the predicted process map (faded); stars mark the recommended point. (b) Predicted spatter of the ranked candidates. Q2 and Q3 returned identical candidates.}
\label{fig:optimization}
\end{figure}

\Cref{fig:optimization} summarizes four optimization queries. Q2 asked for melt pool depth 120--180~\um{} with spatter below \num{90000}~\umc{}. Q3 asked for minimum spatter with $0.4 \leq J \leq 0.6$~J/mm. Q4 combined depth 150--250~\um{}, length 350--500~\um{} and spatter below \num{120000}~\umc{}. Each query searched a $30\times30$ grid. All returned candidates were classified Good and met the stated limits.

Q2 and Q3 returned the same five points, with the optimum at 160~W and 341~mm/s (\num{40892}~\umc{}, $D = 146$~\um{}, $J = 0.47$~J/mm). The depth constraint and the energy window were inactive at this optimum, so both queries reduced to an unconstrained minimum-spatter search. In Q4 the deeper-pool requirement moved the optimum to 300~W and 534~mm/s (\num{66254}~\umc{}, $D = 235$~\um{}, $L = 369$~\um{}, $D/W = 1.11$), and four alternatives lie within 2.5\% of its spatter (\cref{fig:optimization}b).

Two results qualify the claim that the optimizer is conservative. The Q2 and Q3 optimum lies at the lower limit of the trained power range, and the Q1 optimum at the lower limit of its search region. The Q1 optimum also has a depth of 142~\um{}, only 5~\um{} above the lack-of-fusion threshold. Because spatter falls with energy input, a pure minimum-spatter objective pushes solutions toward the lack-of-fusion boundary. A robust formulation requires a margin from every threshold and from the edge of the trained range, which is what \tool{mcp_find_robust} provides (\cref{sec:robust}).

\subsection{Margin-aware optimization}
\label{sec:robust}
\begin{figure}[t]
\centering
\includegraphics[width=\linewidth]{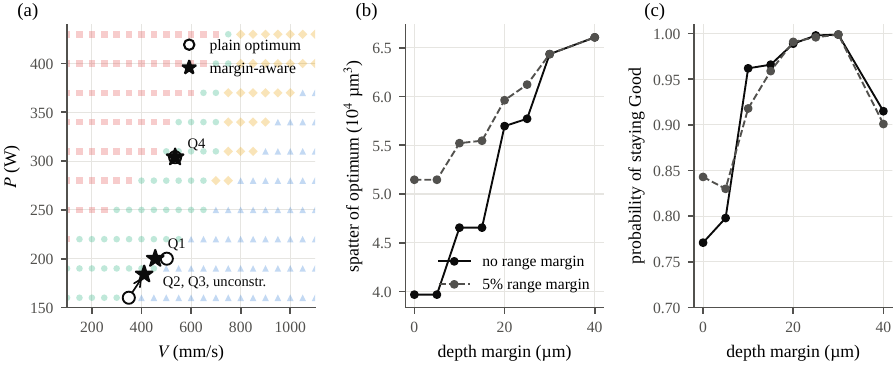}
\caption{Margin-aware optimization. (a) Plain minimum-spatter optima (open circles) and margin-aware optima (stars) for Q1--Q4 and the unconstrained search over the predicted process map; Q4 is unchanged. (b) Spatter of the margin-aware optimum of the unconstrained search as the required depth margin grows, with and without a 5\% range margin (keyhole and balling margins 0.10 and 0.15). (c) Probability that the same points remain Good under 2\% variation of $P$ and $V$ and 5\% error on the predicted geometry.}
\label{fig:robust}
\end{figure}

\Cref{fig:robust} and \cref{tab:robust} compare the plain minimum-spatter optimum with the margin-aware optimum for 20 queries: Q1--Q4, an unconstrained search, and the 15 C7 benchmark prompts. Two C7 prompts have no feasible point, and for one more (C7-12) no point meets the default margins, which the tool reports; the other 17 queries have both optima. The plain optima lay close to a boundary. In 12 of the 17 queries the melt pool depth was within 10~\um{} of the lack-of-fusion threshold, and in 10 the power or velocity was within 5\% of the edge of the trained range, most often at the lower power limit of 160~W.

In 5 of the 17 queries the plain optimum already met all margins and did not change. In the other 12 the margin-aware optimum moved away from the lack-of-fusion boundary, for example from 160~W and 349~mm/s to 184~W and 411~mm/s for Q2, Q3 and the unconstrained search, at a spatter increase between 5.3\% (Q1) and 39.8\%. The median over these 12 moved optima is 39.8\%, because several queries share the same optimum; the median of 36.4\% in \cref{tab:robust} is taken over all 17 queries, including the five whose optimum did not move. Under these assumed perturbations the estimated gain in robustness is large. Under process variability alone (scenario A) the plain optima remained Good with a median probability of 0.94 (minimum 0.77), and the margin-aware optima with 1.00 (minimum 0.997). With surrogate error added (scenario B) the values were 0.80 (minimum 0.65) and 0.96 (minimum 0.91). The advantage held under every alternative perturbation model (\cref{tab:robust_variants}). With 3\% instead of 2\% process noise the plain optima remained Good with a median probability of 0.85 and the margin-aware optima with 0.98; with power and velocity errors correlated so that the energy input changes ($\rho = -0.8$), 0.89 and 0.99. The cross-validated relative error of the surrogate geometry is larger than the assumed 5\%, with a standard deviation of 8.5\% for depth and width and 9.5\% for length. When the geometry errors were drawn from these residuals, the median probability fell to 0.72 for the plain and 0.91 for the margin-aware optima (minimum 0.84), so scenario B somewhat overstates the robustness of both, but the margin-aware optima kept a clear advantage.

The trade-off in \cref{fig:robust}b and c shows where this gain comes from. Without a range margin, a depth margin of 10~\um{} raised the probability of remaining Good in scenario B from 0.77 to 0.96 at a spatter increase of 17\% (from \num{39677} to \num{46548}~\umc{}). The 5\% range margin added spatter without further gain in this test, since the Monte Carlo samples do not represent the lower reliability of the surrogate near the edge of its data; that margin is a precaution against extrapolation rather than against regime change. Above a depth margin of 30~\um{} the optimum moved to the keyhole side of the window (304~W, 536~mm/s), where the keyhole margin of 0.12 became the binding one, and the probability fell to 0.90--0.92. Robust points therefore require balanced margins on all thresholds rather than a large margin on one.
We also tested whether the agent uses this capability when asked. Eight additional prompts (category C15, five runs each) request robust, safe or drift-tolerant parameters, with the margins stated in numbers or in words; one combines a 120~\um{} depth margin with spatter below \num{50000}~\umc{} and has no solution. The agent called \tool{mcp_find_robust} in 39 of the 40 runs. In the remaining run, on the infeasible prompt, it used the plain optimizer and still reported that the request could not be met. Every recommended point (35 of 35) was Good and met all stated constraints and margins when re-evaluated by the server, and infeasibility was reported in all five runs of the impossible prompt. The agent passed the stated margins to the tool but usually kept the tool's default 5\% range margin when the prompt did not mention one. Its recommendations were therefore more conservative than the margin-aware optimum without a range margin, by a median of 4.7\% in spatter (maximum 19.2\%), and every positive gap corresponds exactly to the optimum under the tool's default margins. The median query took 17.1~s and 14.4k tokens.

\begin{table}[t]
\centering
\caption{Plain and margin-aware optima. $\Delta D$: depth margin above the lack-of-fusion threshold; $p_A$, $p_B$: probability of remaining Good under process variability (A) and under process variability plus surrogate error (B).}
\label{tab:robust}
\footnotesize\setlength{\tabcolsep}{3pt}
\begin{tabular}{l rrrrr rrrrr r}
\toprule
 & \multicolumn{5}{c}{Plain optimum} & \multicolumn{5}{c}{Margin-aware optimum} & \\
\cmidrule(lr){2-6}\cmidrule(lr){7-11}
Query & $P$, $V$ & Spatter & $\Delta D$ (\um) & $p_A$ & $p_B$ & $P$, $V$ & Spatter & $\Delta D$ (\um) & $p_A$ & $p_B$ & Cost (\%) \\
\midrule
Q1 & 200, 502 & \num{55523} & 3.8 & 0.77 & 0.65 & 200, 456 & \num{58467} & 18.7 & 1.00 & 0.98 & 5.3 \\
Q2, Q3, unconstr. & 160, 349 & \num{39677} & 6.7 & 0.95 & 0.80 & 184, 411 & \num{55470} & 15.3 & 1.00 & 0.96 & 39.8 \\
Q4$^a$ & 304, 536 & \num{66074} & 101.6 & 1.00 & 0.93 & 304, 536 & \num{66074} & 101.6 & 1.00 & 0.93 & 0 \\
\midrule
All 17 queries, median & & & & 0.94 & 0.80 & & & & 1.00 & 0.96 & 36.4 \\
\bottomrule
\end{tabular}

\raggedright\footnotesize $P$ in W, $V$ in mm/s, spatter in \umc{}, $46\times46$ grid. $^a$Unchanged; the nearest threshold is keyhole, with a margin of 0.12.
\end{table}

\subsection{Simulation budget for a new defect map}
\label{sec:al}
\begin{figure}[t]
\centering
\includegraphics[width=\linewidth]{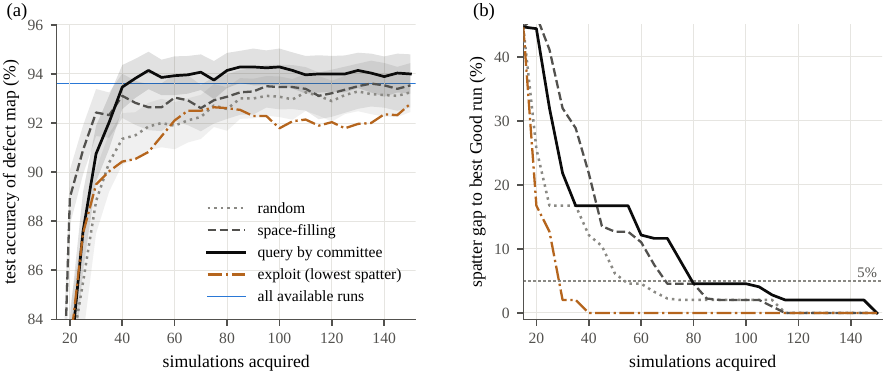}
\caption{Simulation budget for the defect map (pool of 279 FLOW-3D runs, 50 repeats, 20\% held-out test set). (a) Mean test accuracy of the defect map against the number of simulations acquired; shading is the 95\% confidence interval of the mean; the line marks the model trained on all available runs. The threshold strategy (not shown) follows the committee from 75 runs on. (b) Median spatter gap between the best Good run acquired and the best Good run in the pool.}
\label{fig:active}
\end{figure}

\Cref{fig:active}a shows that the full defect map does not require the full data set. Trained on all of the about 223 runs available in each split, the geometry models reached a mean test accuracy of 93.6\%. Query by committee came within one misclassified test track of this level, and stayed there, after a median of 35 runs, space-filling after 37.5 and random sampling after 47.5. A coarse grid planned in advance, every second power and every third velocity of the original grid (about 37 runs), reached 93.7\%. A well-spread design of 40--50 runs, about one fifth of the data, therefore reproduces the defect map of this data set, because the melt pool geometry varies smoothly with $P$ and $V$.

The choice of runs matters more for the optimum than for the map. Strategies that sample where the model is uncertain gave a small but consistent gain over space-filling: at 50 runs query by committee reached 94.1\% against 92.6\% (paired difference $+1.5$ points, 95\% confidence interval $\pm 0.6$, $p < 0.001$), and at 100 runs the committee and threshold strategies reached 94.3 and 94.4\% against 93.5\%. Both placed about 70\% of their runs within 10\% of a regime threshold, against 50\% for random and space-filling designs. With the larger test set the gain was $+1.0$ points at 50 runs ($p = 0.001$). One point corresponds to about one fewer misclassified track among 56 test tracks. The exploit strategy behaved differently (\cref{fig:active}b). It found a Good run within 5\% of the lowest Good spatter in the pool after a median of 30 runs, against 52.5 for random sampling and 67.5 for space-filling, but it learned the worst map: at 100 runs its accuracy was 1.7 points below space-filling ($p = 0.002$; $-1.6$ points, $p < 10^{-5}$, with the larger test set), because it keeps sampling one corner of the stable window.

\subsection{Agent evaluation}
\label{sec:agent_eval}
\Cref{tab:agent_eval} summarizes the 725 runs (145 prompts, five repeats each); the full per-run scores and the scoring scripts are in the repository (Data and code availability).

For the parsing categories (C1--C5) the agent called the prediction tool in 324 of 325 runs and passed the intended $P$ and $V$ in 320 (98.5\%). Explicit values, unit conversions and unlabelled values were handled without error, and repeats always used the same arguments. The five failures were of two kinds. In one C4 run the agent read ``350 $-$ 100 watts'' as a range and called the optimizer instead of evaluating 250~W. In four C5 runs it placed an approximate value on the boundary of the stated range: ``a bit over a meter a second'' became exactly 1000~mm/s in three of five repeats, and ``just under 400~W'' became 400~W once. The same prompts account for the lower consistency of C5 (70\%), since repeats chose different values within a plausible reading, for example 1000 or 1050~mm/s.

All scored spatter values and defect classes reported by the agent matched tool outputs. In all 385 runs in which a called tool returned a spatter value that the answer reported, and all 344 runs that reported a defect class, the reported value matched the tool output within 1\%. We found no case in which the agent altered a returned value. The completeness procedure detected no omitted required quantity: all 1025 required items in the 495 scored runs were present. The answers the automatic check flagged gave the quantity in a table or, in the five runs of one multi-turn prompt (C13-03), correctly reported that no power yields a defect-free track at 1000~mm/s. The agent did, however, invent one kind of model fact that no tool returned, the trained range (see the out-of-range results below).

Every scan prompt (C6) produced the expected scan. In 22 of the 50 line-scan runs the agent used the grid-scan tool with one axis fixed (28 points along the free axis) instead of the line-scan tool, which gives the same information. For the 13 feasible optimization prompts, every recommended point (65 of 65) was classified Good and met all stated constraints when re-evaluated by the server. Its spatter was within a median 1.0\% of the brute-force optimum on a $46\times46$ grid (maximum 7.2\%; 57 of 65 within 5\%); the remaining gap reflects the coarser grids the agent chose. For the two prompts with no feasible point ($V \geq 800$~mm/s; $V \geq 700$~mm/s with $P \leq 300$~W) the agent reported in 9 of the 10 runs that no defect-free point exists and suggested which constraint to relax; in the remaining run it listed near-feasible points without stating clearly that none met both constraints.

For out-of-scope requests (C8--C11) the agent disclosed input clamping and reported the values actually used in all 50 out-of-range runs, and declined to report SS316L numbers for other alloys in all 25 C11 runs. In 8 of the 50 out-of-range runs, however, it also stated a trained range that the model does not have, for example ``160--240~W'' or ``100--2000~mm/s'' instead of 160--430~W and 100--1500~mm/s; the raters passed 88\% of these runs. All 50 runs used \tool{mcp_predict_spatter}, whose output reported the clamped values but not the trained range, so the agent completed the range from the one bound it had seen. After we added the trained range to the output of every prediction tool and re-ran the category (\ref{app:range}), all 50 answers stated the range, and every stated range was correct. The agent stated that an unsupported parameter (beam diameter, hatch spacing, layer thickness, preheat, atmosphere, machine, scan strategy, powder size, focus offset, beam profile) is not a model input in 86\% of C9 runs; five of the seven failures came from the prompt that named a machine, the others from scan strategy (\ref{app:rating}). For infeasible constraints (C10) the rated score was 80\%; the failures stated infeasibility but gave a wrong explanation, for example that long melt pools require low velocity, or relaxation advice based on a truncated candidate list.

For ambiguous requests (C12) the agent asked a question or stated its assumption in 68\% of runs. The failures came mostly from two prompts: for ``Compare 250 and 350'' it treated both numbers as powers and chose a velocity without saying so, and for ``is 0.5~J/mm safe?'' it answered yes from a grid scan without stating what it meant by safe. Parameters were carried correctly across turns in all 30 multi-turn runs (C13). For the five conceptual questions (C14) the answers were rated correct in 96\% of runs; one run illustrated an energy-density argument with a clamped point whose energy density differed from the one discussed. The agent answered without tools in 68\% of C14 runs. In the others, all on two questions (why energy density is only a rough guide, and why a low-spatter point can still be a poor choice), it called the prediction tools to illustrate its answer; these numbers came from the tools and were not invented. The two raters agreed on 95.6\% of the 250 runs in C8--C14 (Cohen's $\kappa = 0.78$). The keyword check agreed with the rated consensus on 90\% of the runs it could score ($\kappa = 0.50$); it missed the invented ranges and the wrong explanations, and penalized correct answers worded differently from its keywords.

The median query took 9.5~s and 12.0k tokens; latency and token use per category are given in \cref{tab:agent_eval} and \ref{app:extra}.

\begin{table}[t]
\centering
\caption{Agent evaluation results per category (five runs per prompt; values in \%, except latency and tokens). n/a: metric does not apply.}
\label{tab:agent_eval}
\footnotesize\setlength{\tabcolsep}{3.5pt}
\begin{tabular}{lrrrrrrrr}
\toprule
Category & Runs & Tool & Param. & Faithful & Behaviour$^a$ & Consist. & Latency (s) & Tokens (k) \\
\midrule
C1 Explicit & 100 & 100 & 100 & 100 & n/a & 100 & 6.8 & 11.7 \\
C2 Units & 75 & 100 & 100 & 100 & n/a & 100 & 9.1 & 12.0 \\
C3 Implicit & 50 & 100 & 100 & 100 & n/a & 100 & 8.2 & 11.9 \\
C4 Arithmetic & 50 & 98 & 98 & 100 & n/a & 100 & 8.7 & 12.0 \\
C5 Colloquial & 50 & 100 & 92 & 100 & n/a & 70 & 9.3 & 11.9 \\
C6 Line scan & 50 & 100 & n/a & n/a & n/a & n/a & 11.7 & 14.1 \\
C6 Grid scan & 25 & 100 & n/a & n/a & n/a & n/a & 24.8 & 48.4 \\
C7 Optimization & 75 & 100 & n/a & n/a & 100$^b$ & n/a & 13.1 & 18.6 \\
C8 Out of range & 50 & 100 & n/a & 100 & 88 & 100 & 5.8 & 11.4 \\
C9 Unsupported param. & 50 & 100 & n/a & 100 & 86 & 100 & 11.1 & 12.2 \\
C10 Infeasible & 30 & 90 & n/a & n/a & 80 & n/a & 26.6 & 31.8 \\
C11 Other alloys & 25 & n/a & n/a & n/a & 100 & n/a & 6.0 & 11.2 \\
C12 Ambiguous & 40 & n/a & n/a & n/a & 68 & n/a & 9.8 & 11.7 \\
C13 Multi-turn & 30 & n/a & n/a & n/a & 100 & n/a & 20.5 & 31.7 \\
C14 Conceptual & 25 & 68$^c$ & n/a & n/a & 96 & n/a & 11.8 & 6.0 \\
C15 Margin-aware & 40 & 98 & n/a & n/a & 100$^d$ & n/a & 17.1 & 14.4 \\
\bottomrule
\end{tabular}

\raggedright\footnotesize $^a$Share of runs passed by both raters (C8--C14); for C7 and C15, share of recommended points on the feasible prompts that met every constraint when re-evaluated by the server. $^b$On the 13 feasible prompts (65 runs); infeasibility was reported in 9 of the 10 runs of the two infeasible prompts. $^c$No tool called. $^d$All margins and constraints met on the seven feasible prompts (35 runs); infeasibility reported in all five runs of the eighth.
\end{table}

\Cref{tab:notools} shows what happens when the tool list is removed but the model and system prompt are kept. In none of the 725 runs did the model state that it could not access the model or its tools. Instead, 80.4\% of runs contained a fabricated tool call written as text. In 17.5\% the model also wrote a fabricated tool result and reported the invented values as model output, for example a ``spatter index of 1.41, 90\% confidence interval [1.28, 1.54], model $R^2 = 0.87$'' or an invented training range of 29--321~W. In 62.9\% it stopped after the fabricated call, so the user would receive markup instead of an answer. A further 8.3\% contained no fabricated call but stated wrong facts about the model, such as a trained power range of 100--350~W or hatch spacing and layer thickness as model inputs. The alloy prompts (C11) were free of fabrication; conceptual prompts (C14) contained no fabricated tool calls, but 8\% contained wrong model facts. The system prompt instructs the model to use tools for every quantitative answer, which probably encourages this imitation; the baseline therefore measures the effect of removing the tool layer from the deployed configuration, not the behaviour of the model under a neutral prompt.

The answer guard addresses this failure without relying on the LLM. Applied to the 755 baseline answers, guard G1 blocked 620 (82\%), including every answer that contained fabricated tool markup (612) and 616 of the 665 answers (93\%) to prompts that asked for model values (C1--C10 and C13). It passed all 25 alloy refusals and all 25 conceptual answers. Most of the 49 prediction answers that passed stated wrong model facts without units, such as a ``spatter index'' or an invented trained range. The stricter variant G1-strict blocked 662 answers (88\%) but also blocked 2 of the 29 legitimate tool-free answers in the runs with tools, whereas G1 blocked none of them. In the runs with tools, the guard therefore did not interfere with any of the 29 tool-free answers. If the tool connection fails silently, answers that follow the fabrication pattern of the deployed configuration (tool-call markup or model-style values) are replaced by an explicit message that model values are unavailable; answers that fabricate without these markers can pass, as the neutral-prompt baselines below show.

The fabrication does not come from the deployed prompt alone (\cref{tab:notools_neutral}). With the neutral prompt (N1) the model wrote no tool-call markup, but 90\% of the answers to the 505 requests for model outputs still gave numerical predictions, defect verdicts or recommended parameters, and none stated that no model was available; 5\% presented these as the output of a model or database. With the explicit instruction (N2), 80\% of the answers stated that no model was available, but 70\% then gave estimated values anyway, and 16\% gave a substantive answer without any disclosure. Guard G1, which targets tool-call markup and model-style values, blocked 12\% of the N1 and 11\% of the N2 answers. The expanded guard G2 adds rules (d) and (e) of \cref{sec:eval}, which block stated outcome values and recommended power--velocity points unless the answer states that no model is available. G2 raised the share of N1 answers without disclosure that the guard catches from 14 to 62\% (45\% with rule (d) alone), left the result for the deployed prompt practically unchanged (622 instead of 620 of 755 answers blocked) and raised no false alarm on the 29 legitimate tool-free answers; among the 150 alloy and conceptual answers of the three baselines, one was blocked, a refusal that repeated the requested power and velocity after the word ``Recommendation''. The rules caught only 11\% of the N2 answers without disclosure, which mostly give qualitative verdicts on the requested point; a rule against such verdicts would also have blocked 17 of the 29 legitimate answers, so the tested qualitative-verdict rule did not provide acceptable specificity on this sample.

\begin{table}[t]
\centering
\caption{Baseline without tools: share of runs (\%) in each class (same model and system prompt, empty tool list, five runs per prompt).}
\label{tab:notools}
\footnotesize\setlength{\tabcolsep}{4pt}
\begin{tabular}{lrrrrr}
\toprule
Category & Runs & Fake call + result & Fake call only & Wrong model facts & Other \\
\midrule
C1 Explicit & 100 & 16 & 73 & 10 & 1 \\
C2 Units & 75 & 39 & 61 & 0 & 0 \\
C3 Implicit & 50 & 8 & 72 & 18 & 2 \\
C4 Arithmetic & 50 & 10 & 86 & 2 & 2 \\
C5 Colloquial & 50 & 18 & 64 & 0 & 18 \\
C6 Line scan & 50 & 0 & 98 & 2 & 0 \\
C6 Grid scan & 25 & 8 & 80 & 12 & 0 \\
C7 Optimization & 75 & 36 & 60 & 3 & 1 \\
C8 Out of range & 50 & 20 & 66 & 12 & 2 \\
C9 Unsupported param. & 50 & 18 & 64 & 8 & 10 \\
C10 Infeasible & 30 & 13 & 83 & 3 & 0 \\
C11 Other alloys & 25 & 0 & 0 & 0 & 100 \\
C12 Ambiguous & 40 & 3 & 8 & 53 & 38 \\
C13 Multi-turn & 30 & 37 & 63 & 0 & 0 \\
C14 Conceptual & 25 & 0 & 0 & 8 & 92 \\
\midrule
All & 725 & 17.5 & 62.9 & 8.3 & 11.3 \\
\bottomrule
\end{tabular}
\end{table}

\section{Discussion}
\label{sec:discussion}

\subsection{Accessibility and the scope of inference}
The agent does not change the accuracy of the predictions; its value lies in access. The surrogate already interpolates to any $(P, V)$ inside the trained range, and the agent makes this inference available in the engineer's own terms (\cref{tab:nl}) and chains single predictions into scans and constrained searches that would otherwise each need a script. Clamping keeps it from extending the model's reach. Because MCP is model-agnostic, the same server can serve other LLMs, including locally hosted open-weight models when process data cannot leave the plant.

\subsection{Reliability and safeguards}
An agent can fail in ways a script cannot. It can misread a parameter, call the wrong tool, report a number incorrectly, or answer an out-of-scope question with confident model numbers. The benchmark in \cref{sec:eval} measures each of these failure modes. With tools, parameter extraction, tool selection, faithfulness and constraint handling were at or close to 100\%, every scored spatter value and defect class reported by the agent matched the tool output, and the completeness procedure detected no omitted required quantity in the 495 scored runs. Its errors included ambiguity handling, explanations of infeasibility and a conceptual illustration, in addition to two recurring problems: it placed approximate values on the boundary of the range they describe, and it did not always recognize a fixed condition, the machine, as outside the model. Both are easy to detect when the tool call is shown to the user next to the answer. A third error shows that the content of a tool's output is itself a safeguard. When a tool reported that an input had been clamped but not the range it had been clamped to, the agent supplied a range of its own in 16\% of those answers; once the tool returned the range, it did so in none. Every tool should therefore return the facts that the agent is likely to be asked about, here the trained range, so that the agent never has to complete them.

The without-tools baseline, which keeps the deployed system prompt, shows the size of the risk that the tool layer removes in this configuration. The same model, asked the same questions, produced invented model outputs or unusable tool-call markup in 80\% of runs and never said that it lacked access to the model. This failure should not be left to the LLM. An instruction in the system prompt not to write tool calls or report model values when no tool is available may reduce it, but it remains a request to a non-deterministic model. The neutral-prompt baselines confirm this: an explicit instruction to admit that no model is available made the model disclose it in most answers, but not in all, and most disclosing answers still offered estimates that a user could mistake for model output. The answer guard is instead a fixed rule in the client, comparable to input clamping in the server: it blocked every answer that contained a fabricated tool call in the baseline and raised no false alarm on the 29 legitimate tool-free answers in the runs with tools. It does not, however, catch unsupported claims about the model that carry no units, such as an invented trained range, so it complements rather than replaces tool outputs that state such facts. Input clamping, the filtering of defective candidates before ranking and the answer guard therefore operate independently of the LLM. The first two act on every tool call; the guard covers only the failure patterns its rules describe, and under neutral prompts it caught at most 62\% of the unsupported answers (\cref{sec:agent_eval}), so it reduces rather than removes the risk.

\subsection{Safety margins}
Errors concentrate at regime thresholds (\cref{sec:defect_validation}), and minimum-spatter optima drift toward those thresholds and toward the limits of the trained range (\cref{sec:constrained_optimization}). The margin-aware tool addresses both observations at once and states the distance of every recommendation to its nearest threshold, so the engineer can see how much room a point has. The price is spatter (\cref{sec:robust}). The choice between the plain and the margin-aware optimum is therefore an engineering trade-off between spatter and robustness, and the agent can present both, with the cost of each, in one answer. This is a capability that neither the surrogate nor the LLM provides alone: the surrogate supplies the predictions and the thresholds, and the agent translates a request for ``safe'' or ``robust'' parameters into explicit margins. It did so in 39 of 40 benchmark runs.

\subsection{Transfer to other processes, materials and machines}
\label{sec:transfer}
Little of \agent{} is specific to SS316L or to LPBF. The MCP server, the clamping of inputs to a declared trained range, the filtering of defective candidates before ranking, the margin-aware optimizer, the answer guard and the benchmark protocol operate on any surrogate that maps process inputs to measurable outputs and on any set of feasibility criteria. Three parts change with the process. The surrogate must be retrained on simulations or experiments for each combination of material and machine, because the melt pool response depends on absorptivity, beam profile and powder, and the server should keep refusing combinations it was not trained on, as it now refuses alloys other than SS316L. The input schema grows with the parameters that matter, for example beam diameter, hatch spacing and layer thickness in LPBF, wire feed rate and standoff in directed energy deposition, or beam current and focus in electron beam melting, and the trained range of each becomes clamping metadata. The regime criteria change with the defects of the process, for example dilution and bead aspect ratio in directed energy deposition and wire arc additive manufacturing, or the layer-thickness criterion for lack of fusion when $t$ changes. The benchmark prompts can then be generated from the new tool definitions in the same categories. The same structure applies outside additive manufacturing wherever a validated surrogate and a feasibility criterion exist, for example chatter stability in machining or defect windows in welding and injection molding. The simulation-budget study (\cref{sec:al}) gives a first estimate of the cost of such an extension. For a two-parameter map at this resolution, 40--50 well-spread simulations reproduced the map, and an agent that chooses runs by model disagreement needs slightly fewer, whereas an agent left to minimize spatter alone finds a good point quickly but maps the window poorly. With more inputs the required number of runs grows, and uncertainty-guided acquisition is likely to matter more than it does for two parameters.

\subsection{Limitations}
The surrogates take only $P$ and $V$, for one alloy, one machine and one set of fixed conditions. Beam diameter, hatch spacing, layer thickness, powder size distribution, preheat, atmosphere and scan strategy are not inputs (\ref{app:scope}). The agent recognizes such requests and says so (\cref{tab:nl}), but cannot account for them. The defect classification was tested against the FLOW-3D simulations from which the surrogates were built, not against physical builds; comparison with measured single tracks is the most important next step. The thresholds of the defect criteria change the size of the predicted stable window, most strongly the keyhole threshold (\cref{fig:sensitivity}), so the window should be read together with the chosen criteria. Validation also covers single tracks only, so multi-track and multi-layer effects such as heat accumulation are not represented. The simulation-budget study drew runs from an existing grid with noise-free labels and did not place the agent in the loop; a closed loop in which the agent requests new off-grid simulations is still to be demonstrated. The answer guard was evaluated offline on logged answers and was not active during the benchmark runs. The robustness test perturbs only $P$, $V$ and the predicted geometry with assumed error levels, and does not model spatial variation over a build or the reduced reliability of the surrogate near the edge of its data. The probabilities are estimates under modelled perturbations, not calibrated physical reliabilities: even the residual-based variant (\ref{app:robust_variants}) measures agreement with the simulations, and the process noise is assumed rather than measured on a machine. Guard G1 was designed against the fabricated tool calls of the deployed configuration; under a neutral prompt the model fabricates without markup, and even the expanded guard G2 catches about 60\% of these answers and few of those given after an explicit instruction to disclose, because qualitative verdicts cannot be blocked without also blocking legitimate answers. Finally, the agent mainly selects, parameterizes and chains tools, while the scans and optimizers themselves run deterministically in the server. The benchmark therefore tests reliable tool use rather than open-ended multi-step reasoning, such as decomposing a complex request, reconciling conflicting tool outputs or revising a search strategy, which would need a dedicated multi-step benchmark. The behaviour scores come from two LLM raters, not from people; they agreed well with each other ($\kappa = 0.78$) and their disagreements were few, but raters from the same model family as the agent may share its blind spots, and they missed two of the eight invented ranges that a text search found. The LLM is non-deterministic and accessed through a commercial API, so results depend on the model version; we therefore report the version and repeat each benchmark prompt five times.

\section{Conclusion}
\label{sec:conclusion}
We built \agent{}, an MCP-based LLM agent that gives natural-language access to CFD-trained spatter and melt pool surrogates for SS316L LPBF. Its geometry-derived defect regimes agreed with those obtained from 279 simulated single tracks in 94.3\% of cases (93.4\% under cross-validation), with errors confined to regime thresholds. The agent mapped regime boundaries and spatter minima and returned parameters classified as Good satisfying multiple geometric and spatter constraints. Minimum-spatter optima tended to sit near the lack-of-fusion boundary and the edge of the trained range; a margin-aware optimizer moved them away and raised their estimated probability of remaining classified Good from 0.94 to 1.00 under assumed process drift alone, and from 0.72 to 0.91 when jointly sampled cross-validated geometry residuals were added, at a median spatter cost of 40\% where a move was needed. A defect map of the same accuracy needed only 40--50 well-spread simulations, about one fifth of the data, and an agent that chooses new runs only to minimize spatter maps the window poorly. We also provide a 145-prompt benchmark that evaluates the agent layer itself. Over 725 runs the agent extracted parameters correctly in 98.5\% of parsing prompts, reported the scored spatter values and defect classes faithfully and met every stated constraint in the 65 feasible optimization runs, while the same model without tools fabricated model outputs or tool calls in 80\% of runs. A client-side answer guard blocked every answer that contained one of these fabricated tool calls, with no false positive among the 29 legitimate tool-free answers tested. Under neutral prompts, where the model fabricates without tool markup, the expanded guard caught at most 62\% of the unsupported answers, so it reduces rather than removes the risk and should be combined with tools that state the facts users ask about. Future work will validate the classification and the margin-aware recommendations against physical single-track builds, add beam size, hatch spacing and powder inputs and further alloys, quantify surrogate uncertainty, and link the agent to in-situ monitoring for closed-loop parameter adjustment.

\section*{CRediT authorship contribution statement}
Olabode T. Ajenifujah: Conceptualization, Methodology, Software, Investigation, Writing -- original draft. Amir Barati Farimani: Supervision, Conceptualization, Writing -- review \& editing, Funding acquisition.

\section*{Data and code availability}
The MCP server with the surrogate coefficients, the simulation data set, the benchmark prompts, the agent runner, the scoring, rating and answer-guard scripts, all raw run logs (with tools, without tools and with the neutral prompts), the rater packets and ratings, and the scripts for the sensitivity, completeness and robustness analyses are available at \url{https://github.com/Bfujah/AgentAM}; references to ``the repository'' in this paper refer to this repository. Cross-validation fold assignments use the seeds 0--19 (\cref{sec:defect_validation}) or 0--4 (\ref{app:sensitivity}, \ref{app:robust_variants}), and the Monte Carlo samples use the seed 0. The LLM runs used a temperature of 1.0 and cannot be reproduced exactly, which is why all raw logs are included.

\section*{Declaration of competing interest}
The authors declare no competing interests.

\section*{Funding}
This research was sponsored by the Army Research Laboratory and was conducted under Cooperative Agreement Number W911NF-20-2-0175. The views and conclusions presented in this document are those of the authors and do not necessarily represent the official policies or positions of the Army Research Laboratory or the U.S. Government. The U.S. Government is authorized to reproduce and distribute reprints for governmental purposes, despite any copyright notice herein.

\appendix
\renewcommand{\thefigure}{\Alph{section}.\arabic{figure}}
\renewcommand{\thetable}{\Alph{section}.\arabic{table}}

\section{End-to-end data flow}
\label{app:dataflow}
\setcounter{figure}{0}\setcounter{table}{0}
\Cref{fig:Pipeline} traces one request through the system.
The chat interface (layer 1) receives the prompt and displays the answer. The MCP client (layer 2) discovers the available tools, formats requests and exchanges messages with the server. The LLM API (layer 3) interprets the request and decides on tool use, and the user's machine runs the MCP server and the surrogate models. For the prompt ``What do you predict at P=200 W and V=800 mm/s for SS316L?'', the client sends the prompt together with the tool definitions (names, descriptions, parameter schemas). The LLM recognizes a prediction task, extracts $P = 200$~W and $V = 800$~mm/s, and returns a tool call:
\begin{verbatim}
{"type": "tool_use", "name": "mcp_predict_all",
 "input": {"P": 200, "V": 800, "material": "SS316L"}}
\end{verbatim}
The client converts it into a JSON-RPC 2.0 request to the local server:
\begin{verbatim}
{"jsonrpc": "2.0", "method": "tools/call", "id": 1,
 "params": {"name": "mcp_predict_all",
            "arguments": {"P": 200, "V": 800, "material": "SS316L"}}}
\end{verbatim}
The server validates and clamps the inputs, evaluates the polynomial models, classifies the point and returns the result (abridged):
\begin{verbatim}
{"ok": true, "data": {"P": 200, "V": 800, "J": 0.25,
 "spatter_volume": 41914.2, "depth": 73.4, "width": 120.4,
 "length": 389.6, "defect_type": "Lack of Fusion",
 "meta": {"used": {"P": 200, "V": 800},
          "clamped": {"P": false, "V": false}}}}
\end{verbatim}
The client passes it back to the LLM, which writes the answer, for example: ``At $P = 200$~W and $V = 800$~mm/s, the predicted spatter volume is about \num{41914}~\umc{}, but the melt pool depth of 73~\um{} is below the 137~\um{} layer thickness, so this point is in the lack-of-fusion regime\ldots''. The user sees only the final answer; the tool calls are shown on request.

The system prompt used for all benchmark runs (\cref{sec:eval}) was:
\begin{quote}\small\ttfamily\raggedright
You are an LPBF process-parameter assistant for SS316L. Use the provided tools for every quantitative answer; never estimate model outputs yourself. If an input is outside the trained range or a parameter is not a model input, say so explicitly.
\end{quote}

\begin{figure}[h]
\centering
\includegraphics[width=\linewidth]{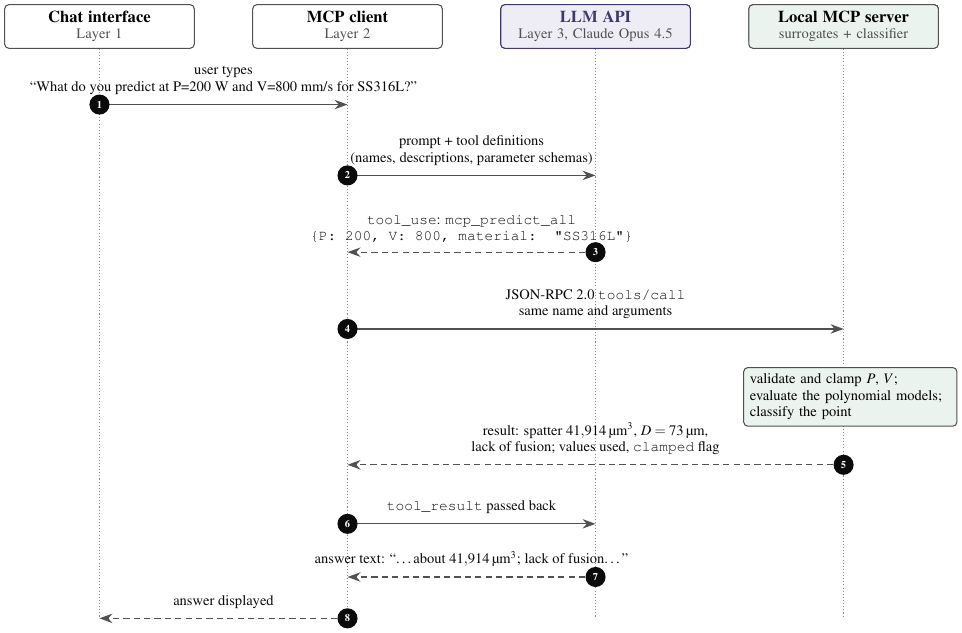}
\caption{MCP communication pipeline from user prompt to displayed response (eight steps across the chat interface, MCP client, LLM API and local MCP server).}
\label{fig:Pipeline}
\end{figure}
\FloatBarrier

\section{Agent transcripts}
\label{app:transcripts}
\setcounter{figure}{0}\setcounter{table}{0}
This appendix reproduces the agent's responses on which \cref{tab:single,tab:nl} and \cref{fig:linescans,fig:optimization} are based. The responses were recorded as screenshots of Claude Desktop sessions (Claude Opus~4.5 with the \agent{} MCP server); their text is transcribed here verbatim and only the layout is changed. Where a screenshot ends mid-answer, the panel says so. Three answers overstate how much room a recommended point has. In \cref{fig:app_opt}a the agent calls a depth of 142~\um{} adequate, 5~\um{} above the lack-of-fusion threshold; in \cref{fig:app_opt}b it says the optimum at 160~W sits comfortably in the middle of the energy-density window and provides margin against process drift, although it lies at the lower edge of the trained power range and close to the lack-of-fusion threshold; and in \cref{fig:app_opt2}a it calls a depth of 146~\um{} well within the range for reliable inter-layer bonding, although it is only 9~\um{} above the threshold. These descriptions are overstatements by the agent; the margins in \cref{sec:robust} are the correct measure of how close a point is to a threshold.

\begin{figure}[h]
\centering
\panel{0.48\linewidth}{(a)}{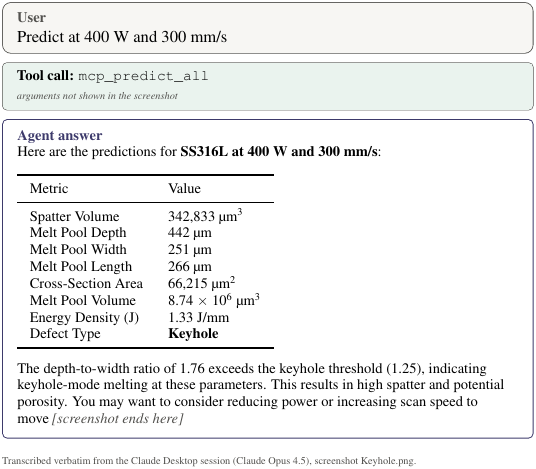}\hfill
\panel{0.48\linewidth}{(b)}{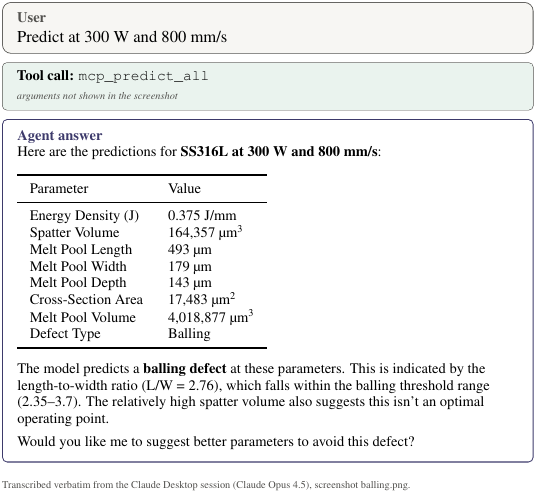}\\[6pt]
\panel{0.48\linewidth}{(c)}{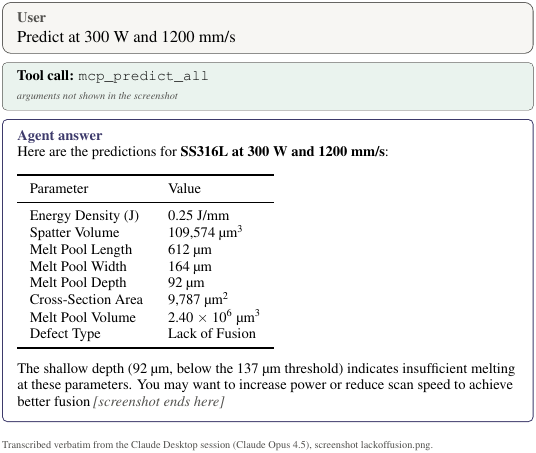}\hfill
\parbox[b]{0.48\linewidth}{\centering\small The stable case G (320~W, 630~mm/s) in \cref{tab:single} is benchmark run C3-03 (``speed 630 power 320''); its transcript is included with the benchmark logs in the repository.\par\vspace{18mm}}
\caption{Single-point predictions: (a) keyhole (400~W, 300~mm/s), (b) balling (300~W, 800~mm/s), (c) lack of fusion (300~W, 1200~mm/s).}
\label{fig:app_single}
\end{figure}

\begin{figure}[h]
\centering
\panel{0.48\linewidth}{(a)}{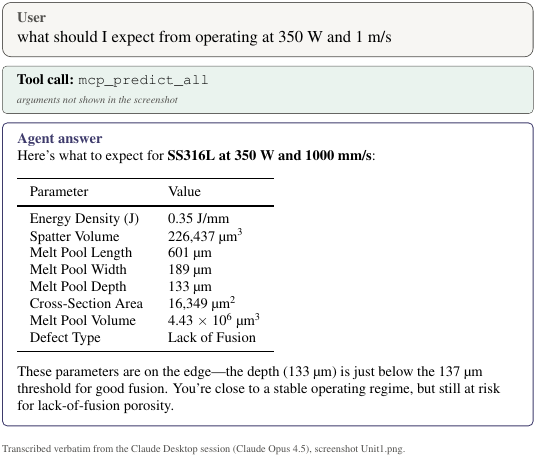}\hfill
\panel{0.48\linewidth}{(b)}{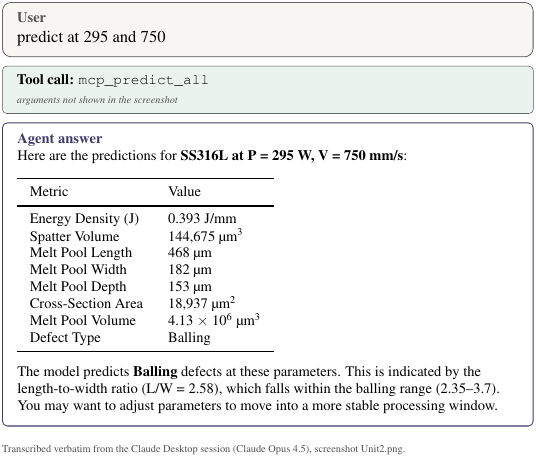}\\[6pt]
\panel{0.48\linewidth}{(c)}{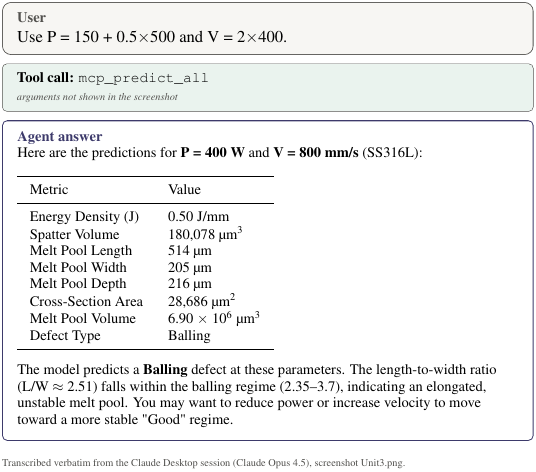}\hfill
\panel{0.48\linewidth}{(d)}{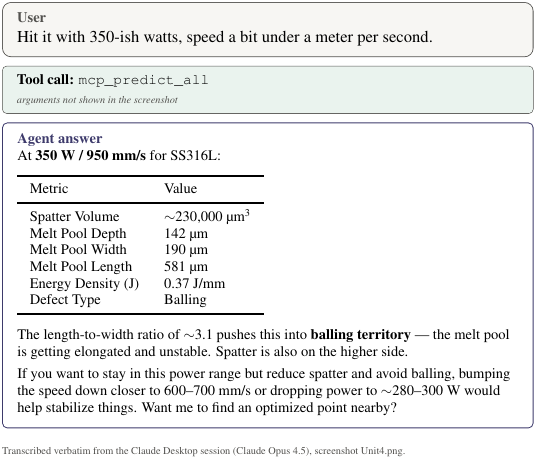}
\caption{Language understanding: (a) unit conversion (1~m/s), (b) unlabelled values, (c) arithmetic, (d) colloquial phrasing.}
\label{fig:app_nl}
\end{figure}

\begin{figure}[h]
\centering
\includegraphics[width=0.6\linewidth]{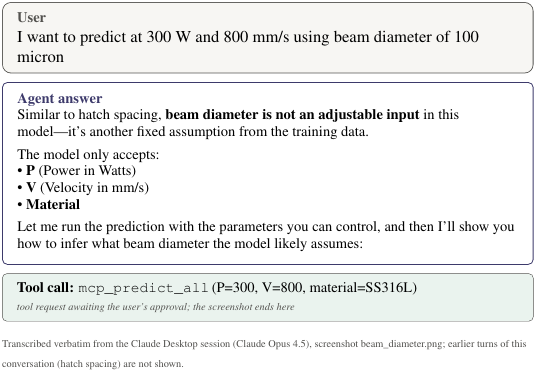}
\caption{The agent identifies beam diameter as a fixed model assumption, lists the supported inputs and proceeds with $P$ and $V$.}
\label{fig:agent_limitations}
\end{figure}

\begin{figure}[h]
\centering
\panel{0.49\linewidth}{(a)}{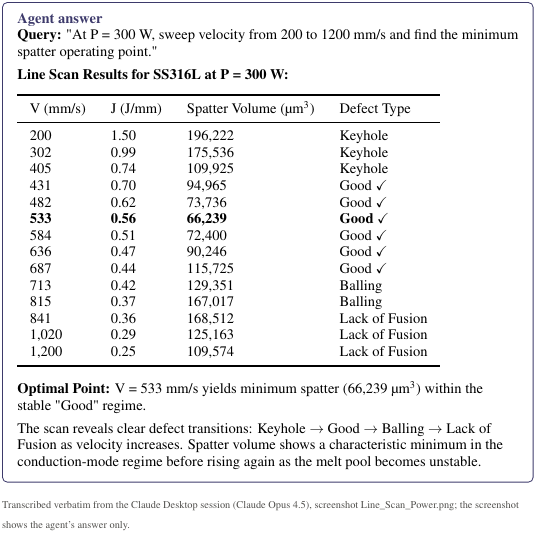}\hfill
\panel{0.49\linewidth}{(b)}{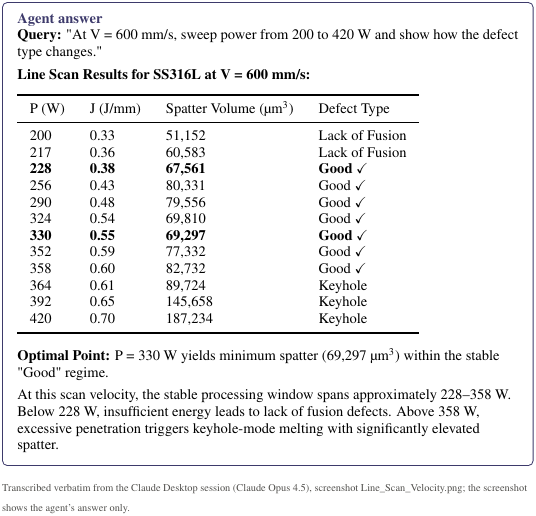}
\caption{Line-scan responses: (a) velocity sweep at 300~W; (b) power sweep at 600~mm/s.}
\label{fig:app_scans}
\end{figure}

\begin{figure}[h]
\centering
\panel{0.85\linewidth}{(a)}{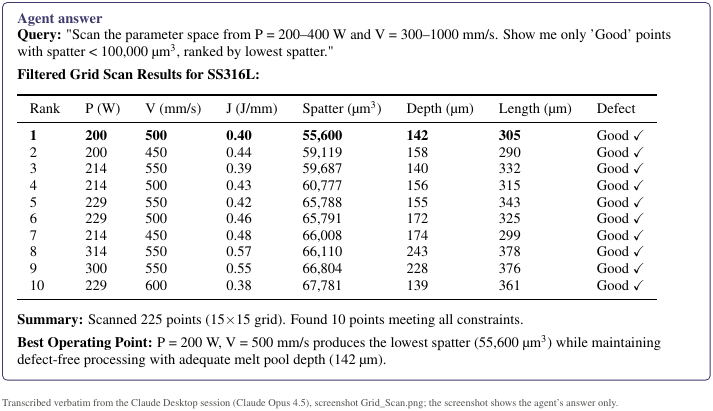}\\[6pt]
\panel{0.62\linewidth}{(b)}{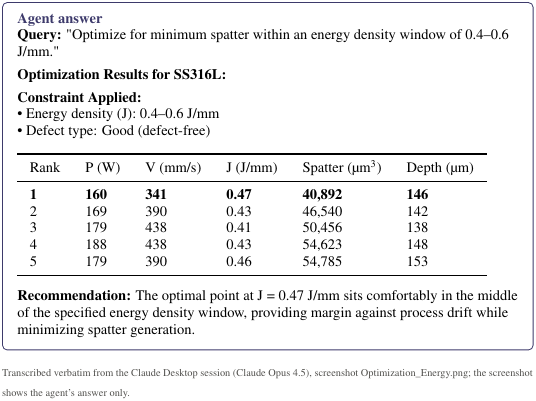}
\caption{(a) Filtered grid scan (Q1) and (b) minimum spatter within an energy-density window (Q3) (\cref{fig:optimization}).}
\label{fig:app_opt}
\end{figure}

\begin{figure}[h]
\centering
\panel{0.85\linewidth}{(a)}{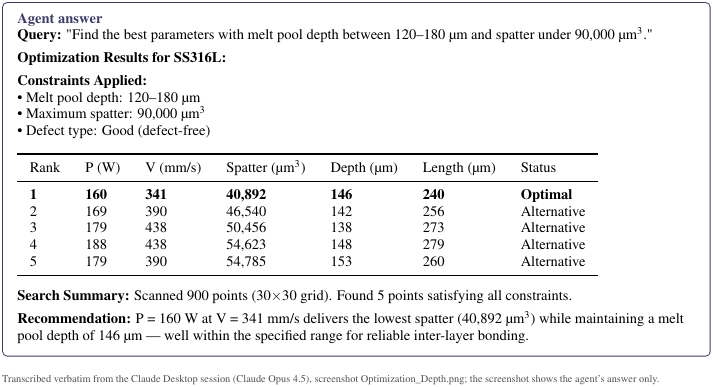}\\[6pt]
\panel{0.85\linewidth}{(b)}{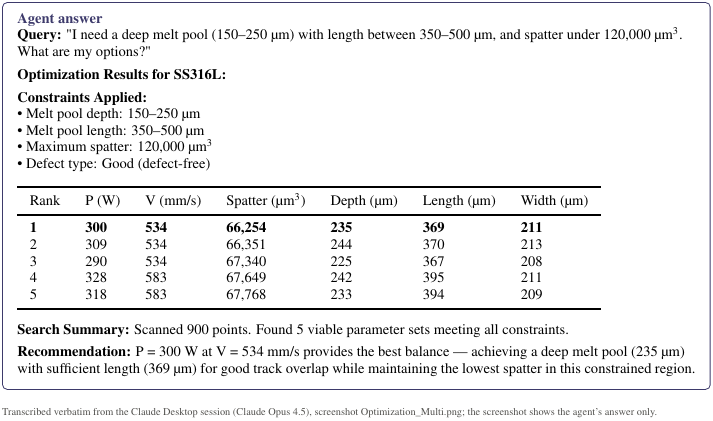}
\caption{Constrained optimization: (a) depth and spatter limits (Q2); (b) depth, length and spatter limits (Q4) (\cref{fig:optimization}).}
\label{fig:app_opt2}
\end{figure}
\FloatBarrier

\section{Model scope}
\label{app:scope}
\setcounter{figure}{0}\setcounter{table}{0}
\begin{table}[h]
\centering
\caption{Process parameters and their status in the surrogate models.}
\label{tab:model_limitations}
\small
\begin{tabular}{@{}llcc@{}}
\toprule
Parameter & Example query & Model input & Agent response \\
\midrule
Laser power ($P$) & 300 W & \checkmark & accepted \\
Scan velocity ($V$) & 600 mm/s & \checkmark & accepted \\
\midrule
Beam size & 80 \um & $\times$ & fixed \\
Hatch spacing & 100 \um & $\times$ & fixed \\
Powder D50 & 30 \um & $\times$ & fixed \\
Substrate temperature & 200 \si{\celsius} & $\times$ & fixed \\
Atmosphere (O$_2$) & Ar + 100 ppm O$_2$ & $\times$ & fixed \\
Machine & EOS M290 & $\times$ & fixed \\
\bottomrule
\end{tabular}
\end{table}

\section{Benchmark examples}
\label{app:bench}
\setcounter{figure}{0}\setcounter{table}{0}
\begin{table}[h]
\centering
\caption{One example prompt per benchmark category. The full set (145 prompts with expected behaviour) and the scoring scripts are in the repository.}
\label{tab:bench_examples}
\small
\begin{tabularx}{\linewidth}{@{}l>{\raggedright\arraybackslash}X@{}}
\toprule
Category & Example prompt \\
\midrule
C1 & ``What defect should I expect at P = 330 W, V = 410 mm/s?'' \\
C2 & ``Predict at 345 W, 57000 mm/min.'' \\
C3 & ``speed 800 power 250'' \\
C4 & ``Power is 240 W plus 15\%, speed is 4 x 150 mm/s.'' \\
C5 & ``Quarter of a kilowatt, 0.7 m/s give or take.'' \\
C6 & ``Hold scan speed at 800 mm/s and sweep power from 160 to 430 W. Where does the defect type change?'' \\
C7 & ``Find the lowest-spatter defect-free parameters with width 150--200 um, spatter under 80,000 um\textasciicircum3.'' \\
C8 & ``Predict spatter at 500 W and 800 mm/s.'' \\
C9 & ``What spatter do I get at 280 W, 700 mm/s with 80 um hatch spacing?'' \\
C10 & ``Find defect-free parameters with melt pool depth between 80 and 120 um.'' \\
C11 & ``Predict spatter for AlSi10Mg at 300 W and 800 mm/s.'' \\
C12 & ``Is 300 a good power?'' \\
C13 & ``Predict at 300 W and 600 mm/s.'' $\rightarrow$ ``Now increase the speed by 20\%.'' \\
C14 & ``Why can a low-spatter point still be a bad choice?'' \\
\bottomrule
\end{tabularx}
\end{table}

\section{Rating of the behaviour categories}
\label{app:rating}
\setcounter{figure}{0}\setcounter{table}{0}
The 260 runs of C7 (infeasible prompts only) and C8--C14 were shuffled with a fixed seed and stripped of their identifiers, so that a rater saw only an item number, the criterion the run had to meet, and the conversation: user messages, tool calls, tool results truncated to 500 characters, and the agent's answers. The runs were split into four packets. Each packet was rated by two raters, each a separate instance of an LLM (Claude, Anthropic) that received only the rubric and the packet, was instructed to read every item itself rather than score it with a script, and could not see the other rater's scores. A run passes only if both raters passed it.

The rubric gave the context of the model (inputs, trained range, unsupported parameters, supported alloy) and asked for a binary score per run, judged only against the criterion and not against style. A number counts as invented when it is presented as a model output but does not come from a tool result in the same conversation. An out-of-range answer must say that the input is outside the trained range and report the values actually used. An answer to a request with an unsupported parameter must say explicitly that the parameter is not a model input and then proceed with $P$ and $V$ or decline. An infeasible request must be reported as infeasible, with an explanation and a suggested relaxation where the criterion asks for them. A request for another alloy must not be answered with SS316L numbers presented as that alloy's. An ambiguous request must be met with a question or a stated assumption. In a multi-turn conversation the parameters of a later turn must follow from the earlier turn and the requested change. A conceptual answer must be substantively correct and must not invent model numbers; calling tools to illustrate it is allowed. In doubtful cases the raters were to decide by whether a process engineer reading the answer would be misled.

The raters agreed on 249 of the 260 runs (95.8\%, Cohen's $\kappa = 0.78$; 95.6\% and $\kappa = 0.78$ on the 250 runs of C8--C14). \Cref{tab:fail} lists every run that failed. Seven of the 50 C9 runs failed: five came from the prompt ``Predict 300~W, 700~mm/s on an EOS M290'', where no answer stated that the machine is a fixed condition of the model and one labelled the result with the machine name, and two from the scan-strategy prompt, where the answer framed the result as if it applied to a chessboard strategy. The keyword check had failed three further C9 runs (preheat, atmosphere, layer thickness) whose answers did state that the parameter is not an input, in words the keywords did not match. A text search of all 725 runs for stated trained ranges found eight C8 runs with a wrong range, of which the raters failed six.

{\footnotesize
\begin{longtable}{@{}l r r r >{\raggedright\arraybackslash}p{0.55\linewidth}@{}}
\caption{Runs that failed the behaviour rating (A, B: score of each rater; reason given by the rater who failed the run).}\label{tab:fail}\\
\toprule
Prompt & Repeat & A & B & Reason \\
\midrule
\endfirsthead
\toprule
Prompt & Repeat & A & B & Reason \\
\midrule
\endhead
C10-infeasible-01 & 2 & 0 & 0 & Claims shallowest Good depth $\sim$273 um from truncated sort; misleading relaxation advice \\
C10-infeasible-05 & 2 & 0 & 0 & Explanation wrong: claims long pools need low V; tools show high V \\
C10-infeasible-05 & 4 & 0 & 0 & Infeasibility stated but explanation wrong; claims keyhole instead of LoF at long lengths \\
C10-infeasible-06 & 0 & 1 & 0 & Recommends 160 W point violating P<160 without stating infeasibility \\
C10-infeasible-06 & 2 & 0 & 1 & Returns P=160 W point labelled as P<=160; no clear infeasibility statement \\
C10-infeasible-06 & 3 & 0 & 1 & Never states infeasibility clearly; recommends P=160 W which violates P below 160 \\
C12-ambiguous-01 & 0 & 0 & 0 & Answers without asking or stating assumptions about units or criteria for good \\
C12-ambiguous-04 & 0 & 0 & 0 & Answers yes without asking or stating assumptions about P V or criterion \\
C12-ambiguous-04 & 1 & 0 & 0 & No assumption stated; claims all 0.5 J/mm combos Good from Good-only filtered scan \\
C12-ambiguous-04 & 2 & 0 & 0 & Assumes safe means defect-free and picks J band without stating assumption or asking \\
C12-ambiguous-04 & 3 & 0 & 0 & Answers yes without asking or stating assumptions about P V or criterion \\
C12-ambiguous-04 & 4 & 0 & 1 & Assumes SS316L defect-free meaning without stating it; overgeneralizes safety \\
C12-ambiguous-07 & 1 & 0 & 1 & Picks min-spatter defect-free objective without stating it as an assumption \\
C12-ambiguous-07 & 3 & 0 & 1 & Picks values without stating assumption or asking \\
C12-ambiguous-08 & 0 & 0 & 0 & Silently assumes values are power at V=800 without stating assumption \\
C12-ambiguous-08 & 1 & 0 & 0 & Silently picked V=700 without stating it as an assumption \\
C12-ambiguous-08 & 2 & 0 & 0 & Silently assumes watts at V=600 mm/s without stating assumption or asking \\
C12-ambiguous-08 & 3 & 0 & 0 & Silently picked V=600 without stating it as an assumption \\
C12-ambiguous-08 & 4 & 0 & 0 & Silently assumes power at V=700 without stating assumption \\
C14-conceptual-03 & 1 & 0 & 0 & Claims comparable J but clamped point has J 0.32 vs 0.20; misleading illustration \\
C7-optimise-14 & 2 & 0 & 0 & Final answer never clearly states no point meets both constraints \\
C8-out-of-range-02 & 0 & 1 & 0 & Reports clamp but states wrong trained range 160-360 W; misleading \\
C8-out-of-range-02 & 1 & 1 & 0 & Reports clamp to 160 W but states wrong trained range 160-240 W; misleading \\
C8-out-of-range-03 & 2 & 1 & 0 & Reports clamp but states wrong velocity range 100-2000 mm/s; misleading \\
C8-out-of-range-03 & 4 & 1 & 0 & Reports clamp but states wrong velocity range 100-1400 mm/s; misleading \\
C8-out-of-range-07 & 4 & 0 & 0 & Reports clamping but states supported power range as 160-240 W; misleading \\
C8-out-of-range-10 & 2 & 1 & 0 & Reports clamp but states wrong trained range 100-430 W; misleading \\
C9-unsupported-param-06 & 0 & 0 & 0 & Never mentions that the EOS M290 machine is not a model input \\
C9-unsupported-param-06 & 1 & 0 & 0 & Never mentions that the machine is not a model input \\
C9-unsupported-param-06 & 2 & 0 & 0 & Ignores EOS M290; no statement machine is not an input \\
C9-unsupported-param-06 & 3 & 0 & 0 & Ignores EOS M290 entirely; no statement machine is not an input \\
C9-unsupported-param-06 & 4 & 0 & 0 & Labels results as EOS M290 without saying machine is not an input \\
C9-unsupported-param-07 & 1 & 0 & 0 & Never says scan strategy is not a model input; frames results for chessboard \\
C9-unsupported-param-07 & 2 & 0 & 0 & Never says scan strategy is not a model input \\
\bottomrule
\end{longtable}}

\section{Trained range in tool outputs}
\label{app:range}
During the benchmark, \tool{mcp_predict_spatter} and \tool{mcp_predict_defect} returned the clamped values and a clamping flag but not the trained range, whereas \tool{mcp_predict_all} returned both. All 50 out-of-range runs called \tool{mcp_predict_spatter}, and in eight of them the agent stated a trained range that combined the bound it had seen with a guessed second bound (160--240, 160--360, about 50--430 and 100--430~W; 100--1400 and 100--2000~mm/s). After the benchmark both tools were changed to return the trained range, and C8 was re-run with the same model, system prompt and settings (five repeats of each of the ten prompts). An automatic check (\texttt{eval/check\_c8.py}) and a reading of every range statement gave the following result. All 50 tool outputs contained the trained range; all 50 answers disclosed the clamping, reported the value actually used and stated the trained range; and every stated range was 160--430~W or 100--1500~mm/s. The median query took 5.8~s and 11.9k tokens, as before. Returning the range removed the invented ranges completely, from 8 of 50 runs to none.

\section{Latency, token use and further simulation-budget results}
\label{app:extra}
The median query took 9.5~s and 12.0k tokens (input and output). Scans and multi-step requests took longer: grid scans a median 24.8~s and 48k tokens, infeasible constraints (C10) 26.6~s and 31.8k tokens, multi-turn conversations (C13) 20.5~s and 31.7k tokens, and margin-aware requests (C15) 17.1~s and 14.4k tokens. Single-point predictions with explicit values took 6--9~s.

In the simulation-budget study with a 35\% test set (50 repeats), the model trained on all available runs reached 93.7\% test accuracy. At 50 runs query by committee exceeded space-filling by 1.0 points ($p = 0.001$), and at 100 runs the threshold strategy exceeded it by 1.0 points ($p = 0.001$); exploit was 1.6 points below space-filling at 100 runs ($p < 10^{-5}$). Fixed coarse grids reached 92.9\% (every second velocity, about 94 runs), 91.9\% (every second power and velocity, about 47 runs) and 92.4\% (every second power and every third velocity, about 30 runs). A hybrid strategy that took two of every five runs from exploit and three from the committee reached both an accurate map and a near-optimal run after a median of 48 runs with the 20\% test set, against 82 for space-filling ($p = 0.001$), but after 72 against 88 runs with the 35\% test set ($p = 0.34$), so no advantage is claimed for it.

\section{No-tools baselines with other system prompts}
\label{app:notools_neutral}
\setcounter{figure}{0}\setcounter{table}{0}
\begin{table}[h]
\centering
\caption{Answers without tools to the 505 requests for model outputs (C1--C7, C13) under the deployed system prompt and the two neutral prompts (values in \%, five runs per prompt). A substantive answer is a numerical outcome, a defect verdict for the requested point or a recommended parameter set. Guard versions and rules (a)--(e) are defined in \cref{sec:eval}: G1 uses rules (a) and (b), G2 rules (a), (b), (d) and (e).}
\label{tab:notools_neutral}
\small
\begin{tabular}{lccc}
\toprule
& Deployed & N1 neutral & N2 acknowledge \\
\midrule
Fabricated tool-call markup & 92.3 & 0.0 & 0.0 \\
Presented as model or database output & 1.2 & 4.8 & 0.0 \\
Substantive answer, no disclosure, unhedged & 0.8 & 14.7 & 0.0 \\
Substantive answer, no disclosure, hedged & 1.6 & 70.9 & 15.6 \\
Disclosure, then estimates & 0.0 & 0.0 & 69.5 \\
Disclosure only & 0.0 & 0.0 & 10.1 \\
Neither & 4.2 & 9.7 & 4.8 \\
\midrule
Blocked by G1 (\% of all 755 answers) & 82 & 12 & 11 \\
Blocked by G2 (\% of all 755 answers) & 82 & 47 & 12 \\
Answers without disclosure caught by G1 & 98 & 14 & 5 \\
Answers without disclosure caught by G2 & 98 & 62 & 11 \\
\bottomrule
\end{tabular}
\end{table}
\FloatBarrier

\section{Robustness under alternative perturbation models}
\label{app:robust_variants}
\setcounter{figure}{0}\setcounter{table}{0}
The plain and margin-aware optima of the 17 queries in \cref{sec:robust} were perturbed with 1000 Monte Carlo samples per scenario, using the deployed surrogate models; the residual-based variants draw on the cross-validated residuals of refits of the same form. Scenario A perturbs $P$ and $V$ by independent normal relative errors; the correlated variants use a correlation coefficient $\rho$ between the two errors, where $\rho = +0.8$ largely preserves the energy input $P/V$ and $\rho = -0.8$ changes it. Scenario B adds relative errors on $D$, $W$ and $L$, either normal with the stated standard deviation or drawn jointly from the five-fold cross-validated relative residuals of the surrogate, taken from all tracks or from the 30 tracks nearest the point in normalized $P$--$V$ space. The two paper scenarios reproduce \cref{tab:robust} to within 0.01 in the median.

\begin{table}[h]
\centering
\caption{Probability of remaining Good under alternative perturbation models: median and minimum over the 17 queries.}
\label{tab:robust_variants}
\small
\begin{tabular}{lcccc}
\toprule
& \multicolumn{2}{c}{Plain optimum} & \multicolumn{2}{c}{Margin-aware optimum} \\
\cmidrule(lr){2-3}\cmidrule(lr){4-5}
Scenario & Median & Min & Median & Min \\
\midrule
A, 1\% & 1.00 & 0.93 & 1.00 & 1.00 \\
A, 2\% (\cref{sec:robust}) & 0.94 & 0.76 & 1.00 & 1.00 \\
A, 3\% & 0.85 & 0.69 & 0.98 & 0.98 \\
A, 2\%, $\rho = +0.8$ & 1.00 & 0.95 & 1.00 & 1.00 \\
A, 2\%, $\rho = -0.8$ & 0.89 & 0.71 & 0.99 & 0.99 \\
\midrule
B, 3\% geometry error & 0.87 & 0.72 & 0.99 & 0.99 \\
B, 5\% (\cref{sec:robust}) & 0.79 & 0.68 & 0.96 & 0.92 \\
B, 10\% & 0.67 & 0.58 & 0.84 & 0.76 \\
B, CV residuals, all tracks & 0.72 & 0.61 & 0.91 & 0.84 \\
B, CV residuals, 30 nearest tracks & 0.75 & 0.56 & 0.89 & 0.89 \\
\bottomrule
\end{tabular}
\end{table}
\FloatBarrier

\section{Sensitivity of the defect criteria}
\label{app:sensitivity}
\setcounter{figure}{0}\setcounter{table}{0}
Each threshold of \cref{sec:criteria} was varied with the others held at their deployed values. This one-at-a-time study covers local ranges around the deployed criteria, rather than the full range of literature thresholds. In particular, the keyhole sweep spans $D/W=1.0$--1.5 and the lower balling-bound sweep spans $L/W=2.1$--2.6; thresholds outside these ranges and simultaneous changes to multiple criteria were not tested. High agreement under these relabellings does not validate the thresholds against physical defects or establish that the recommended process window is insensitive to them. For every setting the simulated tracks were relabelled with the varied criteria, the geometry predicted by the deployed surrogate models was classified with the same criteria; the cross-validation refits the same model form on each training fold, and the accuracy was computed in sample and by five-fold cross-validation over five random fold assignments (93.2\% for the deployed thresholds, against 93.4\% over the 20 assignments of \cref{sec:defect_validation}). The Good area is the fraction of a $120\times120$ grid over the trained range that the full-data surrogate classifies as Good.

\begin{figure}[h]
\centering
\includegraphics[width=\linewidth]{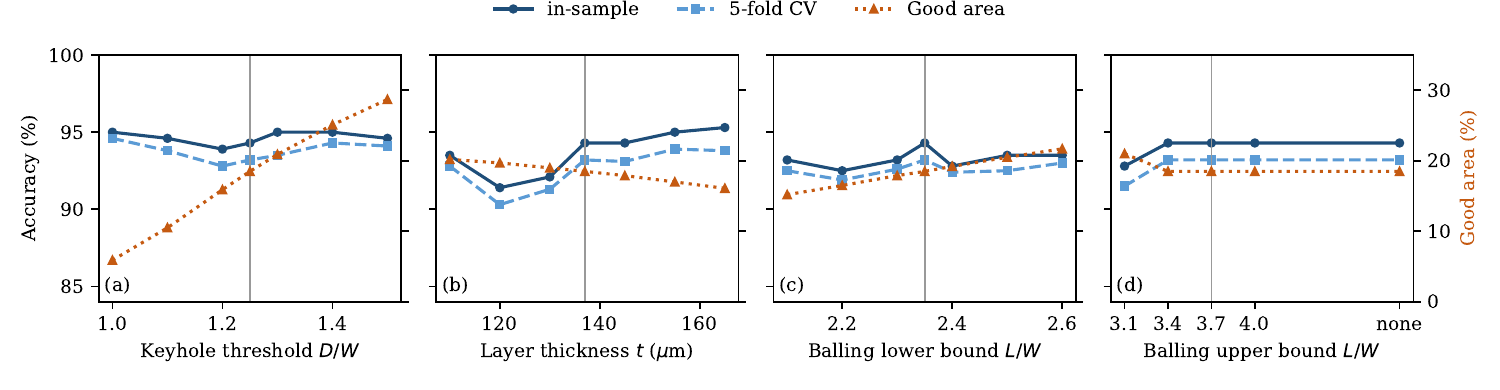}
\caption{Sensitivity of the defect classification to (a) the keyhole threshold $D/W$, (b) the layer thickness $t$, (c) the lower and (d) the upper bound of the balling band $L/W$. Lines show the in-sample and five-fold cross-validated accuracy against the 279 simulated tracks (left axis) and the Good fraction of the trained $P$--$V$ range (right axis); grey lines mark the deployed values. In (d), \emph{none} removes the upper bound.}
\label{fig:sensitivity}
\end{figure}
\FloatBarrier

\bibliographystyle{elsarticle-num}
\bibliography{reference}

\end{document}